\documentclass{article}

    \PassOptionsToPackage{numbers, compress}{natbib}

 \usepackage[preprint]{OFP}

\usepackage[utf8]{inputenc} % allow utf-8 input
\usepackage[T1]{fontenc}    % use 8-bit T1 fonts
\usepackage{hyperref}       % hyperlinks
\usepackage{url}            % simple URL typesetting
\usepackage{booktabs}       % professional-quality tables
\usepackage{amsfonts}       % blackboard math symbols
\usepackage{nicefrac}       % compact symbols for 1/2, etc.
\usepackage{microtype}      % microtypography
\usepackage{xcolor}         % colors

\usepackage{multirow}

\usepackage{float}
\usepackage[ruled,linesnumbered,noline]{algorithm2e}
\SetAlgoCaptionSeparator{~}
\SetAlCapHSkip{0pt}
\SetNlSkip{0.25em}
\makeatletter
\SetNlSty{textnormal}{\footnotesize}{:}
\SetInd{0.2em}{0.41em}
\renewcommand{\@algocf@post@ruled}{%
  \kern\interspacealgoruled
  \hrule width\algocf@ruledwidth height0.4pt\relax
}
\makeatother
\usepackage[table]{xcolor}
\usepackage{wrapfig}
\usepackage{graphicx}
\usepackage{amsmath}
\usepackage{amsthm}
\usepackage{amssymb}
\newtheorem{proposition}{Proposition}
\usepackage{caption}
\usepackage[capitalise]{cleveref}

\usepackage{subcaption}
\usepackage{enumitem}

\title{One-Step Flow Policy: \\ Self-Distillation for Fast Visuomotor Policies}

\author{%
  Shaolong Li \\
  CSE Department, University of Michigan\\
  \texttt{shaolonl@umich.edu}
  \And
  Lichao Sun\footnotemark[2] \\
  Lehigh University\\
  \texttt{lis221@lehigh.edu}
  \And
  Yongchao Chen\footnotemark[2] \\
  College of AI, Tsinghua University\\
  \texttt{yongchaochen12@gmail.com}
}

\begin{document}

\maketitle
\begingroup
\renewcommand{\thefootnote}{\fnsymbol{footnote}}
\footnotetext[2]{Corresponding authors.}
\endgroup

\begin{abstract}
  Generative flow and diffusion models provide the continuous, multimodal action distributions needed for high-precision robotic policies. However, their reliance on iterative sampling introduces severe inference latency, degrading control frequency and harming performance in time-sensitive manipulation. To address this problem, we propose the One-Step Flow Policy (OFP), a from-scratch self-distillation framework for high-fidelity, single-step action generation without a pre-trained teacher. OFP unifies a self-consistency loss to enforce coherent transport across time intervals, and a self-guided regularization to sharpen predictions toward high-density expert modes. In addition, a warm-start mechanism leverages temporal action correlations to minimize the generative transport distance. Evaluations across 56 diverse simulated manipulation tasks demonstrate that a one-step OFP achieves state-of-the-art results, outperforming 100-step diffusion and flow policies while accelerating action generation by over $100\times$. We further integrate OFP into the $\pi_{0.5}$ model on RoboTwin 2.0, where one-step OFP surpasses the original 10-step policy. These results establish OFP as a practical, scalable solution for highly accurate and low-latency robot control.
\end{abstract}

\section{Introduction}
\label{sec:intro}

Vision-Language-Action (VLA) models are progressing quickly in robot manipulation \citep{cheang2024gr,gbagbe2024bi,fan2025interleave,liu2024rdt}, autonomous driving \citep{zhou2025autovla,yuan2025autodrive}, and long-horizon task execution \citep{lin2025onetwovla,zheng2024tracevla,ajay2023compositional}. Within this high-dimensional control setting, flow and diffusion models have emerged as a dominant paradigm for parameterizing conditional policies \citep{black2024pi_0,intelligence2025pi_,intelligence2025pi,liu2024rdt}. These generative models offer two distinct advantages over discrete tokenization: they naturally represent the multimodal action distributions inherent in human demonstrations, and they output continuous control signals. This continuous output is essential for high-precision manipulation tasks where subtle action granularity determines success.

Despite these benefits, the current paradigm suffers from a critical inference latency bottleneck. Sampling from a flow or diffusion policy typically requires iteratively solving an Ordinary Differential Equation (ODE) or a Stochastic Differential Equation (SDE), transporting samples from a noise prior to the target action distribution. This process requires tens to hundreds of forward passes through a large neural network for a single action \citep{song2020denoising,lipman2022flow}. In time-sensitive robotics applications such as high-speed grasping or dynamic interaction, this latency is prohibitive. Action generation delays directly reduce control frequency and exacerbate compounding errors, frequently resulting in task failure. As VLA architectures scale in size and complexity, a central challenge arises:

\textit{How can we accelerate generative policies to output high-fidelity actions in a few steps, or even a single step, without compromising control precision?}

\begin{wrapfigure}{r}{0.45\textwidth}
    \centering
    \vspace{-10pt}
    \includegraphics[width=\linewidth]{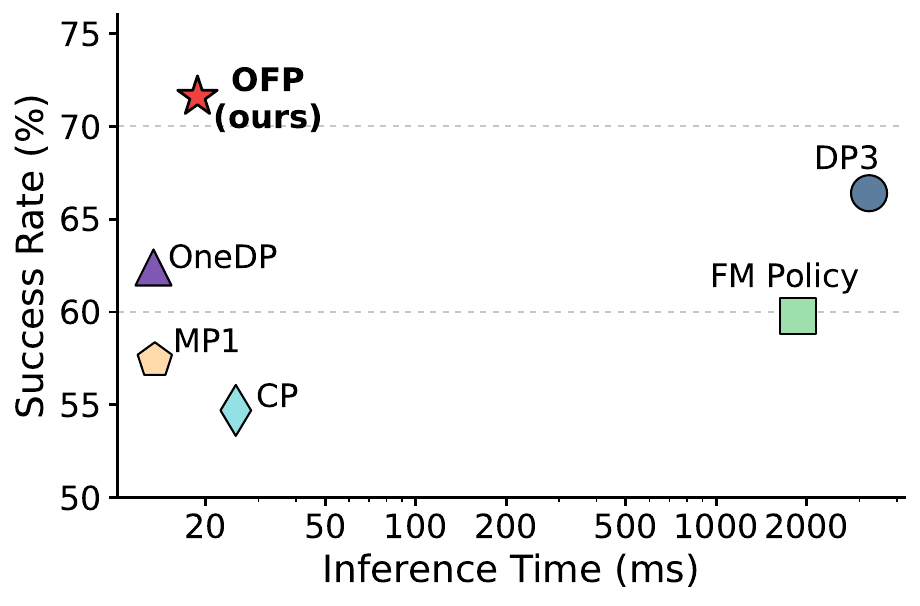} 
    \caption{Averaged across 56 tasks. Evaluated at NFE=1, OFP outperforms all other single-step baselines and accelerates generation by over 100$\times$ compared to DP3 and FM Policy (NFE=100).}
    \vspace{-10pt}
    \label{fig:latency}
\end{wrapfigure}

A large body of work in image and video generation studies how to accelerate flow and diffusion sampling \citep{song2023consistency,kim2023consistency,geng2025mean,yin2024one}, aiming to produce high-fidelity outputs with a minimal Number of Function Evaluations (NFE). We focus on flow-based models due to their widespread industrial deployment \citep{bjorck2025gr00t,black2024pi_0,intelligence2025pi_}. Diffusion models can be viewed as a specific instance of the flow framework, allowing acceleration techniques to transfer across parameterizations. One common approach employs higher-order numerical solvers to reduce steps \citep{lu2022dpm,lu2025dpm}. However, in the extreme few-step regime, discretization errors become unmanageable and severely degrade generation quality.

Another line of work uses distillation \citep{salimans2022progressive}, typically by training a student to compress the trajectory of a pre-trained teacher into fewer steps. Two primary families exist: Consistency Distillation (CD) \citep{song2023consistency,kim2023consistency} and Score Distillation (SD) \citep{yin2024one,yin2024improved,lukoianov2024score}. CD enforces a self-consistency constraint where points along a trajectory map to the same endpoint, enabling one-step generation. SD instead treats the teacher score as a differentiable prior to optimize the generator. Recently, MeanFlow \citep{geng2025mean,geng2025improved} attempted one-step generation by modeling an average velocity field, but it introduces Jacobian-vector products (JVPs) during training, which significantly increase memory costs and destabilizes optimization.

Prior works have directly adapted these vision acceleration ideas into robot policy learning to reduce inference latency, but their impact on control tasks has not been sufficiently analyzed. As a result, existing methods often struggle to achieve both low latency and high action precision. Consistency Policy (CP) \citep{prasad2024consistency} follows the consistency-distillation paradigm. However, due to the mode-covering nature of its trajectory-matching objectives \citep{xu2025one}, the policy tends to average over multimodal action distributions. While this aids few-step inference, the resulting one-step actions often lack the sharpness required for precise manipulation. In contrast, One-Step Diffusion Policy (OneDP) \citep{wang2024one} employs a score-distillation approach. These objectives are typically mode-seeking \citep{lu2025adversarial,wang2024taming}, causing the policy to collapse toward a single high-probability mode. While this produces high-quality one-step samples, it drastically reduces diversity. Due to their design, OneDP-like methods only support single-step inference, preventing them from trading computation time for higher control accuracy.

Motivated by these observations, we propose \textbf{One-Step Flow Policy (OFP)}, which combines the strengths of both directions without requiring an auxiliary network or a pre-trained teacher. OFP is a from-scratch self-distillation framework that unifies two complementary signals: self-consistency, which enforces cross-time coherence of the transport dynamics, and self-guidance, which injects a distribution-level correction that sharpens one-step predictions toward high-density expert behaviors. This combination enables high-fidelity one-step inference without performance degradation. Additionally, we introduce a warm-start action reuse mechanism that provides a strong prior for one-step generation, improving both temporal smoothness and control precision.

% Our contributions are summarized as follows: 1) We present a unified self-distillation approach for flow-based policies that resolves the trade-off between inference speed and action precision without relying on teacher models. 2) We repurpose a warm-start action initialization strategy as a highly effective, training-free mechanism to reduce generative difficulty in few-step inference. 3) We achieve state-of-the-art success rates across 56 simulated tasks spanning Adroit \citep{kumar2016manipulators}, DexArt \citep{bao2023dexart}, and MetaWorld \citep{yu2020meta}. A 1-NFE OFP outperforms 100-NFE diffusion and flow baselines, delivering over a $100\times$ acceleration. 4) We validate the scalability of OFP by integrating it into the $\pi_{0.5}$ model \citep{intelligence2025pi_}. On RoboTwin 2.0 \cite{chen2025robotwin}, our 1-step integration exceeds the baseline 10-step policy, proving its robustness for large-capacity systems and complex semantics.

Our contributions are summarized as follows:
\begin{itemize}[leftmargin=1.5em, itemsep=2pt, topsep=2pt]
    \item We present a unified self-distillation approach for flow-based policies that resolves the trade-off between inference speed and action precision without relying on teacher models.
    \item We repurpose a warm-start action initialization strategy as a highly effective, training-free mechanism to reduce the transport distance in few-step inference.
    \item We achieve state-of-the-art success rates across 56 simulated tasks spanning Adroit \citep{kumar2016manipulators}, DexArt \citep{bao2023dexart}, and MetaWorld \citep{yu2020meta}. A 1-NFE OFP outperforms 100-NFE diffusion and flow baselines, delivering over a $100\times$ acceleration.
    \item We validate the scalability of OFP by integrating it into the $\pi_{0.5}$ model \citep{intelligence2025pi_}. On RoboTwin 2.0 \citep{chen2025robotwin}, our 1-step integration exceeds the baseline 10-step policy, proving its robustness for large-capacity systems and complex semantics.
\end{itemize}

% \vspace{-4pt}
\section{Preliminaries}
\vspace{-4pt}
\subsection{Flow Matching} \label{sec21}
Let the data space be $\mathbb{R}^d$. We aim to learn a time-dependent velocity field $\mathbf{v}_\theta\colon[0,1]\times\mathbb{R}^d\,\mathord{\to}\,\mathbb{R}^d$ defining an Ordinary Differential Equation (ODE). The flow $\{\boldsymbol{\xi}(t)\}_{t\in[0,1]}$ induced by this ODE transports a base distribution $p_0$ to a target data distribution $p_1$:
\vspace{-4pt}
\begin{equation}
 \frac{d}{dt}\boldsymbol{\xi}(t)=\mathbf{v}_\theta\bigl(\boldsymbol{\xi}(t),t\bigr),\quad
 \boldsymbol{\xi}(0)\sim p_0.
 % \tag{2.1} 
 \label{eq21}
\end{equation}
\vspace{-12pt}

Consider a probability path $\{p_t\}_{t\in[0,1]}$ interpolating between the boundary conditions $p_0$ and $p_1$. If $\mathbf{v}_\theta$ matches the vector field generating this path, the solution $\boldsymbol{\xi}(1)$ will follow the target distribution $p_1$. Flow Matching trains $\mathbf{v}_\theta$ by regressing a target velocity field. However, the marginal velocity field is typically intractable.

To make training tractable, Conditional Flow Matching (CFM) \citep{lipman2022flow} introduces a conditional probability path. Given a data sample $x_1\!\sim\! p_1$ and noise $\epsilon\!\sim\! p_0$, we define a conditional path $\mathbf{z}_t\!=\!a(t)x_1 + b(t)\epsilon,\ t\in[0,1],$ where $a(0)\!=\!0,\ b(0)\!=\!1$ and $a(1)\!=\!1,\ b(1)\!=\!0$. The time derivative is given by: $\dot{\mathbf{z}}_t(x_1,\epsilon)=a'(t)\,x_1 + b'(t)\,\epsilon.$ The marginal velocity field corresponds to the expectation: $\bar{\mathbf{v}}(x,t)=\mathbb{E}\!\left[\dot{\mathbf{z}}_t(x_1,\epsilon)\ \middle|\ \mathbf{z}_t=x\right].$ CFM learns $\mathbf{v}_\theta$ by directly regressing the derivative of the conditional path:
 \begin{equation}
 \mathcal{L}_{\mathrm{CFM}}(\theta)
 =\mathbb{E}_{t,x_1,\epsilon}
 \bigl\|\mathbf{v}_\theta(\mathbf{z}_t,t) - \dot{\mathbf{z}}_t(x_1, \epsilon)\bigr\|_2^2.
 % \tag{2.2} 
 \label{eq22}
 \end{equation}
% \vspace{-2pt}
Minimizing \cref{eq22} yields the same optimal solution as regressing the intractable marginal velocity field $\bar{\mathbf{v}}(x,t)$. 

Sampling involves integrating \cref{eq21} from $\mathbf{z}_0\! \sim\! p_0$ using the learned field $\mathbf{v}_\theta$. In practice, this requires time discretization (e.g., Euler or RK4 solvers), resulting in multiple network forward passes. This iterative process is the main inference bottleneck. Since the marginal velocity field exhibits high curvature due to the averaging of overlapping trajectories, standard solvers incur large truncation errors in low-NFE regimes, leading to poor sample quality.

\vspace{-4pt}
\subsection{Flow-Based Generative Policy}
% \vspace{-3pt}
We consider a visuomotor imitation learning setting. Let $\mathcal{D}=\{(o^{(i)}, a^{(i)})\}_{i=1}^N$ denote a dataset of expert demonstrations. The observation $o \in \mathcal{O}$ may include proprioception, multi-view RGB images, pointclouds, and language instructions. The action $a\in\mathbb{R}^{d_a H}$ represents a trajectory chunk of horizon $H$, parameterized typically as end-effector $\mathrm{SE}(3)$ poses or joint angles. The objective is to learn a conditional generative policy $\pi_\theta(a\mid o)$ that samples actions consistent with the expert distribution.

We extend the framework in \cref{sec21} to the conditional setting by defining an observation-conditioned velocity field $\mathbf{v}_\theta(x,t\mid o)$. At deployment, the policy generates an action $\hat{a}$ by integrating the conditional ODE from a noise prior $\mathbf{z}_0 \sim p_0$ to the terminal state $\mathbf{z}_1$. The inference cost scales linearly with NFE.

During training, we employ an Optimal Transport (OT)  \citep{liu2022rectified,liu2022flow} formulation to construct a straight-line conditional path interpolating between noise and data: $\mathbf{z}_t = (1-t)\,\epsilon + t\,a,\ \epsilon\sim p_0.$ The derivative is constant: $\dot{\mathbf{z}}_t = a-\epsilon.$ We train the conditional velocity field via the following regression objective:
\vspace{-3pt}
 \begin{equation}
 \mathcal{L}_{\text{policy}}(\theta)
 =\mathbb{E}_{(o,a)\sim \mathcal{D}}\ \mathbb{E}_{t\sim p_T,\epsilon\sim p_0}
 \bigl\|\mathbf{v}_\theta(\mathbf{z}_t,t\mid o) - (a-\epsilon)\bigr\|_2^2.
 % \tag{2.3}
 \end{equation}
 \vspace{-6pt}
Here $p_T$ is a chosen sampling distribution over $t\in[0,1]$.

% \vspace{-4pt}
\section{One-Step Flow Policy}
\vspace{-4pt}
In this section, we introduce the One-Step Flow Policy (OFP), a from-scratch self-distillation framework designed for fast and high-precision visuomotor control. \cref{sec31} details Self-Consistency Training, which learns an interval-averaged velocity field to enforce temporal coherence across trajectories. \cref{sec:self_guidance} proposes Self-Guided Regularization, a mechanism that explicitly sharpens single-step action predictions toward high-density expert modes. \cref{sec:warm_start} introduces a Warm-Start mechanism leveraging the temporal correlation of consecutive action chunks to provide a strong initialization prior.

\subsection{Self-Consistency Training} \label{sec31}
To eliminate the reliance on iterative ODE integration during inference, we learn an interval-averaged velocity field rather than the instantaneous velocity used in standard Flow Matching \citep{lipman2022flow,liu2022rectified}. Concretely, we parameterize $\mathbf{u}_\theta(\mathbf{z}_t,t,r\mid o)$ for $0\le t \le r \le 1$, representing the average velocity along the trajectory from $t$ to $r$. For brevity, we omit the condition $o$ when the context is clear.

Given $\mathbf{u}$, the update rule for any interval is defined as: $\mathbf{z}_r=\mathbf{z}_t+(r-t)\mathbf{u}(\mathbf{z}_t,t,r).$
 % \begin{equation}
 % \mathbf{z}_r=\mathbf{z}_t+(r-t)\mathbf{u}(\mathbf{z}_t,t,r).
 % \tag{3.1}
 % \end{equation}
When $r\to t^+$, $\mathbf{u}(\mathbf{z}_t,t,r)$ reduces to the instantaneous velocity, maintaining theoretical consistency with Flow Matching on the diagonal.

Given a data-noise pair $((o,a), \epsilon)\!\sim\!\mathcal{D} \!\times\! p_0$, we sample two times $(t,r)\!\sim\! p_{T,R}$ subject to $t\!<\!r$, and draw an intermediate time $m\!\in\![t,r]$. Using the OT straight-line interpolation, we obtain $\mathbf{z}_m\!=\!(1-m)\epsilon+ma$. An Exponential Moving Average (EMA) copy of the model, denoted $\mathbf{u}_{\theta^-}$, serves as the teacher to predict the trajectory endpoint $\hat{\mathbf{z}}_r=\mathbf{z}_m+(r-m)\mathbf{u}_{\theta^-}(\mathbf{z}_m,m,r).$ Based on the definition of average velocity over the interval $[t,r]$, we define the training target:
\vspace{-2pt}
 \begin{equation}
 \mathbf{u}_{\text{target}}=\frac{\hat{\mathbf{z}}_r-\mathbf{z}_t}{r-t}.
 % \tag{3.2}
 \end{equation}
\vspace{-8pt}

\begin{figure}[tb]
    \centering
    \includegraphics[width=\linewidth]{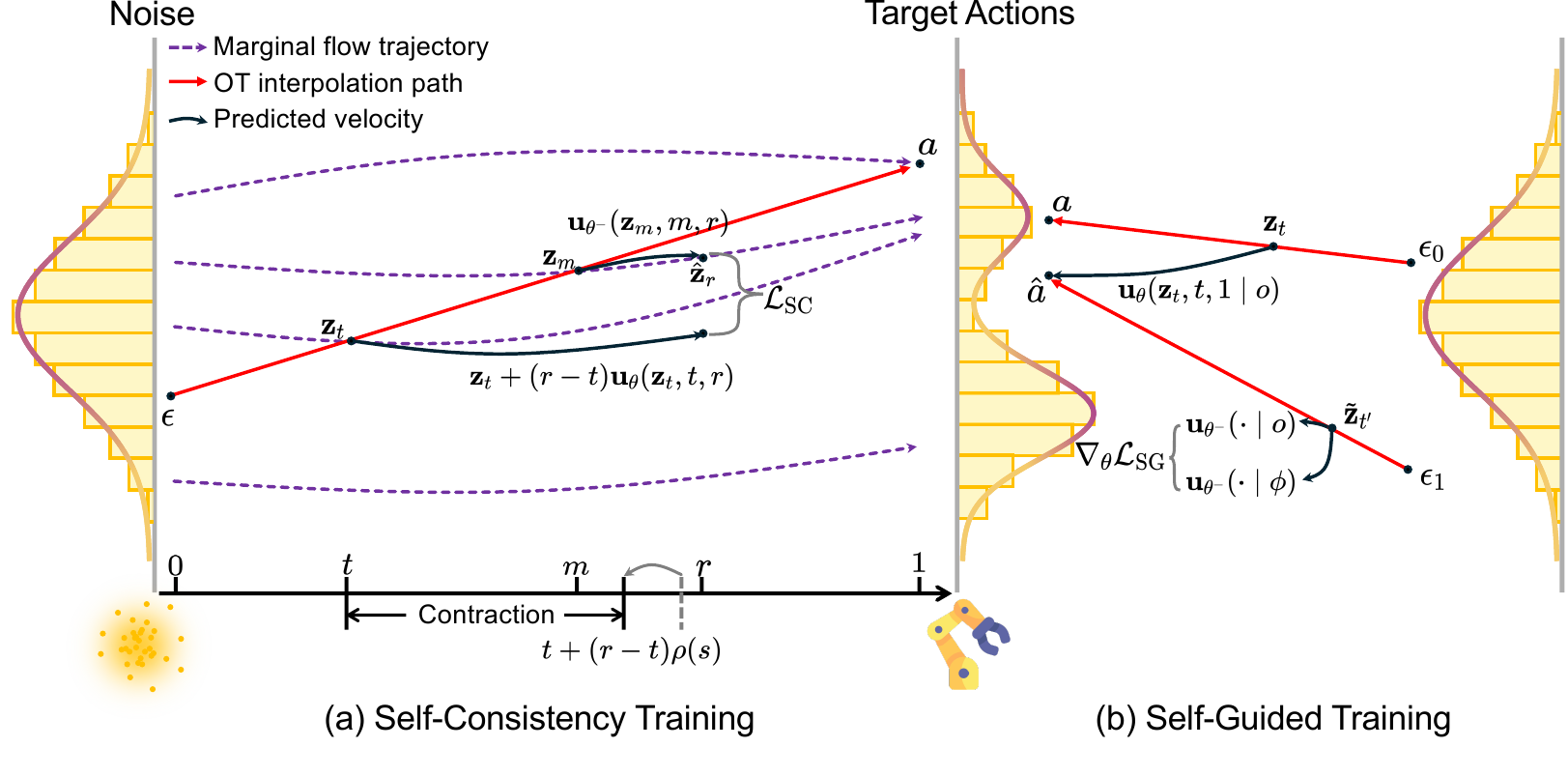} 
    \caption{\textbf{Self-Distillation of One-Step Flow Policies.} (a) \textbf{Self-Consistency Training:} The model learns an interval-averaged velocity field by matching predictions across nested sub-intervals, enforcing temporal coherence along the marginal flow trajectory. (b) \textbf{Self-Guided Training:} By leveraging Classifier-Free Guidance on the model's own predictions, we extract a distribution-level correction signal. The regularization repels single-step predictions from the unconditional prior and sharpens the generated actions toward the high-density modes of the expert data.}
    \label{fig:self_distillation}
    \vspace{-12pt}
\end{figure}

We train $\mathbf{u}_\theta$ by minimizing the self-consistency loss:
\begin{equation}
\begin{split}
&\mathcal{L}_{\text{self-consistency}}(\theta)
= 
\mathbb{E}_{(o,a)\sim \mathcal{D}}\ 
   \mathbb{E}_{(t,r,m)\sim p_{T,R,M},\,\epsilon\sim p_0} 
 \bigl\|\mathbf{u}_\theta(\mathbf{z}_t,t,r \mid o) - \mathbf{u}_{\text{target}}\bigr\|_2^2 .
\end{split}
% \tag{3.3}
\label{eq33}
\end{equation}
 This objective functions as a self-distillation mechanism where the teacher is the slowly evolving EMA copy $\theta^-$.

\textbf{Time-Contracting Schedule}. We sample $m$ from a range that gradually contracts toward $t$ as training proceeds:
 \begin{equation}
 m \sim \mathcal U\Bigl[t, t+(r-t)\rho(s)\Bigr].
 % \tag{3.4}
 \end{equation}
Here, $\rho(s)$ is a monotonically decreasing function of the training step $s$. In the early stages of training ($\rho(s) \approx 1$), $m$ is sampled across the entire interval $[t, r]$. This allows the target $\mathbf{u}_{\text{target}}$ to rely more heavily on the interpolated state $\mathbf{z}_m$, rather than an untrained teacher $\mathbf{u}_{\theta^-}$. This effectively reduces bootstrapping error caused by an untrained teacher. As training progresses ($\rho(s) \to 0$), the sampling range collapses such that $m \to t$. This shifts the objective to enforce strict local self-consistency along the true trajectory.

\begin{proposition} \label{prop31}
Under the assumption that the interval-velocity field is continuously differentiable and globally Lipschitz continuous in state, when the contracting factor $\rho(s) \to 0$ and the EMA teacher $\mathbf u_{\theta^-}$ is accurate, $\mathbf{u}_{\text{target}}$ becomes a consistent supervision signal for the ground truth average velocity $\mathbf{u}(\mathbf{z}_t,t,r)$, and minimizing $\mathcal{L}_{\text{self-consistency}}$ satisfies $\mathbf{u}_\theta\approx \mathbf{u}$.
\end{proposition}
A detailed derivation is provided in Appendix~\ref{app:proof_prop1}. \cref{prop31} establishes that the time-contracting schedule provides a smooth curriculum: it transitions from stabilizing initialization to refining trajectory precision.

\textbf{Comparison with MeanFlow}. Crucially, constructing the target $\mathbf{u}_{\text{target}}$ circumvents the costly Jacobian-Vector Product (JVP) computations required by MeanFlow \citep{geng2025mean,geng2025improved}. MeanFlow relies on a differential objective, which significantly increases memory overhead and wall-clock training time. In contrast, our approach relies solely on forward passes. We provide a formal proof in Appendix~\ref{app:meanflow_equivalence} demonstrating that as the interval $(m-t) \to 0$, our consistency target converges to a finite-difference approximation of the derivative term in the MeanFlow objective.

\textbf{Boundary Anchoring}. To constrain the solution space and ensure the policy remains grounded in the expert data distribution, we introduce a boundary anchoring loss. This term applies standard Flow Matching supervision on the instantaneous velocity:
\begin{equation}\mathcal{L}_{\text{flow}}(\theta) = \mathbb{E}_{(o,a)\sim \mathcal{D}}\ \mathbb{E}_{t\sim\mathcal{U}[0,1],\epsilon\sim p_0} \bigl\|\mathbf{u}_\theta(\mathbf{z}_t,t,t\mid o)-(a-\epsilon)\bigr\|_2^2.
% \tag{3.5}
\label{eq35}
\end{equation}
This anchoring term ensures that OFP retains the capability for high-quality multi-step generation.

The self-consistency objective functions as a trajectory-matching mechanism. This formulation yields strong performance in few-step regimes (e.g., 2 to 4 steps) and preserves the diversity of the action distribution. However, we find that self-consistency alone often yields insufficient precision for complex manipulation in the single-step regime. To address this limitation, we introduce self-guided regularization in \cref{sec:self_guidance}.

\subsection{Self-Guided Regularization}
\label{sec:self_guidance}
While self-consistency (\cref{sec31}) constrains trajectory coherence across time intervals, it does not explicitly enforce that single-step predictions align with the high-density modes of the expert distribution. 

To impose the distribution-level constraint, we introduce a score-based regularizer. The score function $\nabla_x \log p(x)$ points toward higher-density regions. We use it as a guidance signal to steer generated actions toward the expert data manifold.

\textbf{One-step Sample and Re-noising}. Let $\epsilon_0,\epsilon_1\sim \mathcal{N}(0,I)$ be independent noise, and $(o,a)\!\sim\!\mathcal{D}$. For a time $t\in[0,1]$, define the linear mixture $\mathbf{z}_t=(1-t)\epsilon_0+t a.$ The policy generates a one-step prediction by a single interval jump: $\hat{a}\triangleq\hat{\mathbf{z}}_1
=
\mathbf{z}_t+(1-t)\mathbf{u}_\theta(\mathbf{z}_t,t,1\mid o).$ To compare the marginal distribution of the policy with the expert, we re-noise $\hat{a}$ at a new time $t' \sim \mathcal{U}[0,1]$: $\tilde{\mathbf{z}}_{t'} = (1-t')\epsilon_1 + t'\hat{a}.$

Let $\pi_\theta(\tilde{\mathbf{z}}_{t'}\mid o)$ denote the policy-induced marginal distribution of $\tilde{\mathbf{z}}_{t'}$ conditioned on $o$, and let $\pi_\star(\tilde{\mathbf{z}}_{t'}\mid o)$ denote the corresponding expert distribution. We aim to minimize the reverse Kullback-Leibler (KL) divergence between $\pi_\theta(\cdot\mid o)$ and $\pi_\star(\cdot\mid o)$ :
\begin{equation}
 \begin{split}
 &D_{\mathrm{KL}}\big(\pi_\theta(\tilde{\mathbf{z}}_{t'}\mid o)\,\|\,\pi_\star(\tilde{\mathbf{z}}_{t'}\mid o)\big)
 = 
 \mathbb{E}_{\tilde{\mathbf{z}}_{t'}\sim \pi_\theta(\cdot\mid o)}
 \big[\log \pi_\theta(\tilde{\mathbf{z}}_{t'}\mid o)-\log \pi_\star(\tilde{\mathbf{z}}_{t'}\mid o)\big].
 \end{split}
% \tag{3.6}
\label{eq36}
\end{equation}
The gradient of \cref{eq36} with respect to $\theta$ can be written as a score-difference term:
\begin{equation}
 \begin{aligned}
 &\nabla_\theta D_{\mathrm{KL}}(\pi_\theta\,\|\,\pi_\star) \\
 &= 
 \mathbb{E}\Bigl[
 \bigl(\nabla_{\tilde{\mathbf{z}}_{t'}}\log \pi_\theta(\tilde{\mathbf{z}}_{t'}\mid o)
 -\nabla_{\tilde{\mathbf{z}}_{t'}}\log \pi_\star(\tilde{\mathbf{z}}_{t'}\mid o)\bigr)
 \frac{\partial \tilde{\mathbf{z}}_{t'}}{\partial \theta}
 \Bigr] \\
 &=
 \mathbb{E}\Bigl[
 \bigl(s_\theta(\tilde{\mathbf{z}}_{t'}\mid o)-s_\star(\tilde{\mathbf{z}}_{t'}\mid o)\bigr)
 \frac{\partial \tilde{\mathbf{z}}_{t'}}{\partial \theta}
 \Bigr],
 \end{aligned}
 % \tag{3.7}
 \label{eq37}
\end{equation}
where $s(\cdot)=\nabla_{\tilde{\mathbf{z}}_{t'}}\log p(\cdot)$ denotes the score function.

\textbf{Self-Guidance via CFG}. To amplify the conditional signal, we incorporate Classifier-Free Guidance (CFG) \citep{ho2022classifier}. Given a guidance scale $\alpha\ge 1$ and a null condition $\phi$, the CFG-adjusted expert score is defined as: $s_\star^{\mathrm{CFG}}(\tilde{\mathbf{z}}\mid o)
=
s_\star(\tilde{\mathbf{z}}\mid o)
+(\alpha-1)\bigl(s_\star(\tilde{\mathbf{z}}\mid o)-s_\star(\tilde{\mathbf{z}}\mid \phi)\bigr).$ Substituting this form into \cref{eq37} reveals a decomposition into two components:
\begin{equation}
 \begin{aligned}
 &s_\theta(\tilde{\mathbf{z}}_{t'}\mid o)-s_\star^{\mathrm{CFG}}(\tilde{\mathbf{z}}_{t'}\mid o)
 =\\
 &(\alpha\!-\!1)\!\underbrace{\Big(s_\star(\tilde{\mathbf{z}}_{t'}\mid \phi)\!-\!s_\star(\tilde{\mathbf{z}}_{t'}\mid o)\Big)}_{\text{CFG Augmentation}}
 \!+\!\underbrace{\Big(s_\theta(\tilde{\mathbf{z}}_{t'}\mid o)-s_\star(\tilde{\mathbf{z}}_{t'}\mid o)\Big)}_{\text{Distribution Matching}}\!.
 \end{aligned}
 % \tag{3.8}
 \end{equation}
Recent analyses \citep{liu2025decoupled,yu2025self} of DMD-style training suggest that the CFG-Augmentation (CA) term is the primary driver of sample quality. Motivated by this, we use only the CA term as our guidance signal. Intuitively, the CA term steers the model toward the conditional distribution. Unlike prior works \citep{jia2024score,wang2024one} that rely on a pre-trained teacher to compute $s_\star$, we estimate the score using the EMA teacher $\mathbf{u}_{\theta^-}$, which yields a self-guided training signal. 

For the OT probability path $\tilde{\mathbf{z}}_t = (1-t)\epsilon + t a$, the marginal score at time $t$ is analytically related to the instantaneous velocity by:
% \vspace{-8pt}
 \begin{equation}
 s(\tilde{\mathbf{z}}_{t}\mid o)
 =\frac{t\mathbf{u}(\tilde{\mathbf{z}}_{t},t,t \mid o)-\tilde{\mathbf{z}}_{t}}{1-t}.
 % \tag{3.9}
 \end{equation}
 % \vspace{-8pt}
 We then construct the self-guidance target:
\begin{equation}
\begin{split}
 \mathbf s_{\text{target}}=&\mathrm{sg}\Big[\mathbf u_\theta(\mathbf {z}_t,t,1 \mid o)- 
 \big(\mathbf u_{\theta^-}(\tilde{\mathbf{z}}_{t'},t',t' \mid \phi)- \mathbf u_{\theta^-}(\tilde{\mathbf{z}}_{t'},t',t' \mid o)\big)\Big],
 \end{split}
 % \tag{3.10}
\end{equation}
where $\mathrm{sg}[\cdot]$ denotes the stop-gradient operator.

We formulate the self-guidance loss as the squared error between the one-step prediction and the guided target:
\begin{equation}  
\begin{split}
&\mathcal{L}_{\text{self-guidance}}(\theta)
=
\mathbb{E}_{(o,a)\sim \mathcal{D}}\; \mathbb{E}_{t,t'\sim\mathcal{U}[0,1],\,\epsilon\sim p_0}\;\Big[\big\|\mathbf u_\theta(\mathbf {z}_t,t,1 \mid o)-\mathbf s_{\text{target}}\big\|^2\Big]. 
\end{split}
% \tag{3.11}
\label{eq311}
 \end{equation}
Minimizing \cref{eq311} yields a gradient for $\mathbf{u}_\theta$ that aligns the update direction with the CA term (proof provided in Appendix~\ref{app:self_guidance_ca}), which effectively repels the generation from the unconditional mode. Since the guidance signal is induced by the model itself, this objective can also be viewed as a form of self-distillation.

The self-guidance term requires an unconditional branch. We implement this using condition dropout: during training, we randomly replace the observation $o$ with a null token $\phi$ with probability $p_{\text{drop}}$, enabling a single network to capture both conditional and unconditional dynamics.

\textbf{Unified Objective}. We combine the flow anchoring loss (\cref{eq35}), self-consistency loss (\cref{eq33}), and self-guidance loss (\cref{eq311}) into a unified training objective:
\begin{equation}  
\begin{split}
&\mathcal{L}_{\text{self-distill}}(\theta)=
\mathcal{L}_{\text{flow}}(\theta)
 +\lambda_c\mathcal{L}_{\text{self-consistency}}(\theta)
 +\lambda_g\mathcal{L}_{\text{self-guidance}}(\theta).
 \end{split}
 % \tag{3.12}
 \label{eq312}
 \end{equation}

Empirically, adding self-guidance significantly sharpens the precision of one-step outputs. The combination of these signals enables OFP to achieve state-of-the-art performance in both few-step and one-step generation. The complete training procedure is outlined in \cref{alg:ofp_training}.

\textbf{Inference.} OFP supports flexible sampling strategies. For minimal latency, we generate a one-step action prediction $\hat{a}$ via a single global jump from noise: $\hat{a}=\epsilon+\mathbf{u}_\theta(\epsilon,0,1\mid o), \epsilon\sim\mathcal{N}(0,I).$ Alternatively, to achieve higher precision, the model can function as a multi-step solver by chaining interval predictions. Given a time discretization schedule $0=\tau_0 < \tau_1 < \dots < \tau_K=1$, we recursively update the state by querying the average velocity over each sub-interval $[\tau_k, \tau_{k+1}]$: $\hat{\mathbf{z}}_{\tau_{k+1}} = \hat{\mathbf{z}}_{\tau_k} + (\tau_{k+1}-\tau_k)\mathbf{u}_\theta(\hat{\mathbf{z}}_{\tau_k}, \tau_k, \tau_{k+1} \mid o).$ This allows the policy to trade increased inference time for finer control.

% \clearpage
\pagebreak
\subsection{Warm-Start for One-Step Inference}
\label{sec:warm_start}

\begin{wrapfigure}{R}{0.51\textwidth}
\vspace{-40pt}
\setlength{\intextsep}{0pt}
\setlength{\textfloatsep}{0pt}
\begin{algorithm}[H]
\normalsize
\caption{OFP Training}
\label{alg:ofp_training}

\textbf{Inputs:} expert dataset $\mathcal{D}$, model $\mathbf{u}_\theta$, condition dropout $p_{\text{drop}}$.

\textbf{Initialization:} EMA teacher $\theta^- \gets \theta$.

\While{not converged}{
  Sample batch $(o,a)\sim\mathcal{D}$, time variables, and noise.
  
  Apply condition dropout: $o\gets\phi$ with probability $p_{\text{drop}}$.

  Compute $\mathcal{L}_{\text{flow}}$ via \cref{eq35}
  
  Compute $\mathcal{L}_{\text{self-consistency}}$
  via \cref{eq33}
  
  Compute $\mathcal{L}_{\text{self-guidance}}$
  via \cref{eq311}

  Update $\theta$ via \cref{eq312}
  
  Update EMA teacher.
}
\end{algorithm}
\vspace{-20pt}
\end{wrapfigure}

\begin{figure}[tb]
    \centering
    \includegraphics[width=0.85\linewidth]{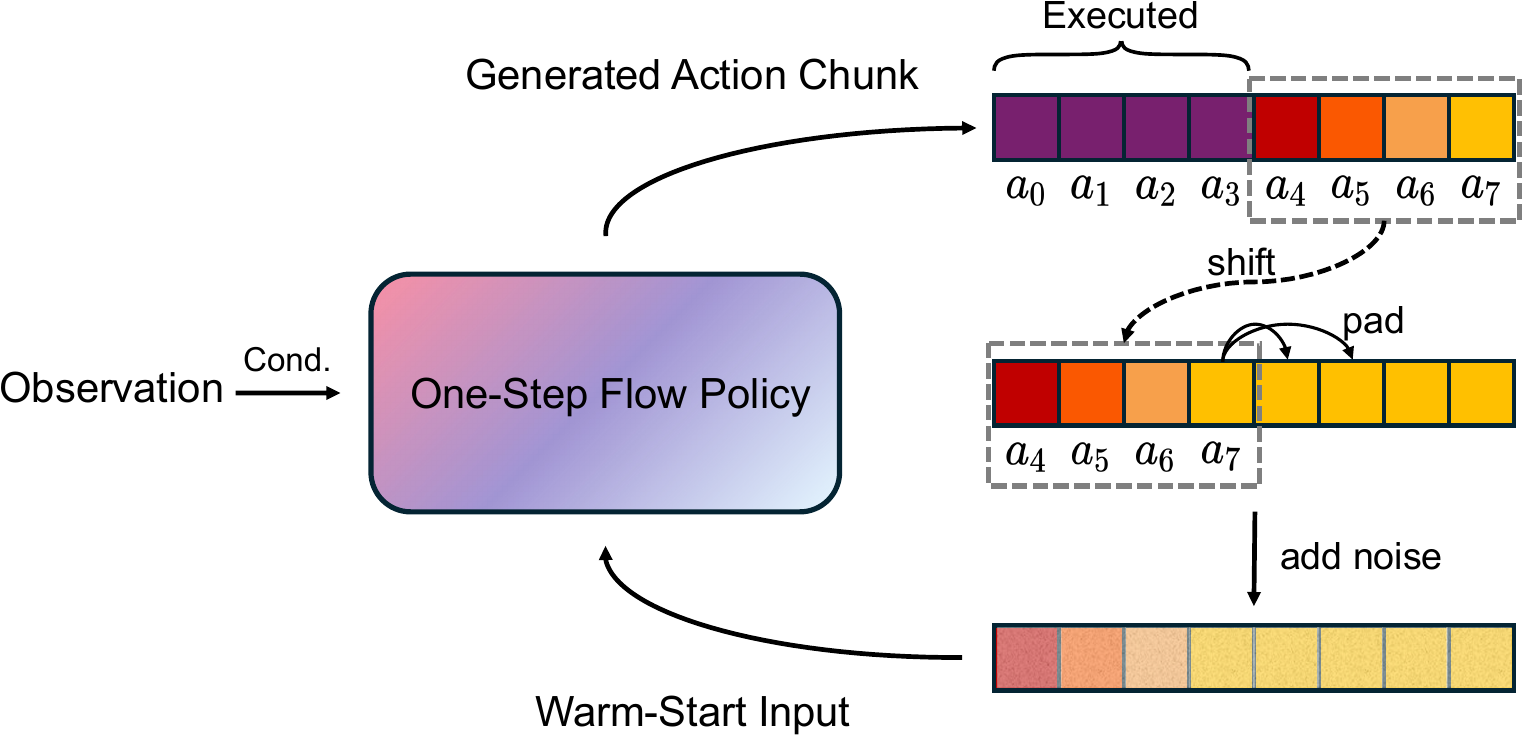} 
    \caption{\textbf{Warm-Start Action Prior for One-Step Inference.} The unexecuted suffix of the previously generated action chunk is shifted and padded with the terminal action to form a full-length temporal prior. By starting closer to the target data manifold rather than from pure Gaussian noise, this initialization reduces the required transport distance.}
    \label{fig:warm_start}
    \vspace{-16pt}
\end{figure}

Warm-start is widely used \citep{chi2025diffusion,janner2022planning} in action-chunking generative policies to improve smoothness and execution efficiency in receding-horizon rollouts. Here, we repurpose warm-start for reducing the generative burden of single-step inference. By leveraging the high temporal correlation between consecutive chunks, we construct a strong prior that lowers the transport cost required to reach the target action distribution.

In a receding-horizon control loop with execution horizon $h < H$, the robot executes the first $h$ actions. The unexecuted suffix represents a valid plan for the current step, $a_{\text{left}}=[a_{h+1},\dots,a_H]\in\mathbb{R}^{d_a (H-h)}.$ We construct a full-length warm-start prior $a_{\text{warm}}$ by shifting the unexecuted suffix and padding the terminal action:

\begin{equation}  a_{\text{warm}}=
\big[a_{h+1},\dots,a_H,\underbrace{a_H,\dots,a_H}_{h\ \text{times}}\big]
 \in\mathbb{R}^{d_a H}.
 % \tag{3.13}
 \end{equation}

\textbf{Warm-Started Generation}. Rather than sampling from non-informative Gaussian noise, we initialize the generator from a noised projection of the warm-start prior. We select a noise level $t_w \in (0, 1]$ and sample $\epsilon \sim \mathcal{N}(0, I)$ to construct the initial state: $\mathbf{z}_{t_w} = (1-t_w)\epsilon + t_w a_{\text{warm}}.$ The policy can then generate the next action chunk $\hat{a}$ via a single interval update: $\hat{a} = \mathbf{z}_{t_w} + (1-t_w)\mathbf{u}_\theta(\mathbf{z}_{t_w}, t_w, 1 \mid o).$ This formulation aligns structurally with the re-noising form used in \cref{sec:self_guidance}, making the inference-time initialization consistent with the training-time guidance construction. We illustrate the construction of this warm-start prior and its integration into the receding-horizon control loop in \cref{fig:warm_start}.

Initializing with $a_{\text{warm}}$ significantly reduces the transport distance the flow must bridge in a single step. By starting closer to the data manifold, the policy yields higher precision and improved temporal smoothness without requiring additional training or computational overhead.

\vspace{-5pt}
\section{Experiments}
\label{sec:experiments}

To evaluate the effectiveness of OFP, we conduct extensive simulation experiments across 4 manipulation benchmarks: Adroit~\citep{kumar2016manipulators}, DexArt~\citep{bao2023dexart}, MetaWorld~\citep{yu2020meta}, and RoboTwin 2.0~\citep{chen2025robotwin}. Our evaluation answers three core questions:
1) Can OFP match or exceed the performance of standard multi-step diffusion and flow policies while requiring only a single network evaluation?
2) How does OFP compare against existing single-step acceleration methods in terms of stability, accuracy, and scaling capability?
3) Does OFP scale effectively to modern VLA architectures without experiencing performance collapse?

\vspace{-5pt}
\subsection{Experimental Setup}
We divide our evaluation into 2D image-based and 3D pointcloud-based control settings. For the 2D evaluation, we test on 3 tasks from Adroit~\citep{kumar2016manipulators} and 4 tasks from DexArt~\citep{bao2023dexart}. For the 3D evaluation, we test on these 7 tasks alongside 49 tasks from MetaWorld~\citep{yu2020meta}. The Adroit and DexArt evaluate single-task learning, while MetaWorld evaluates large-scale multi-task learning. Following prior work~\citep{seo2023masked}, we group the 49 MetaWorld tasks into 4 empirical difficulty levels: Easy, Medium, Hard, and Very Hard.

\textbf{Baselines.} We compare OFP against five strong generative policy baselines: Diffusion Policy (DP)~\citep{chi2025diffusion} and Flow Matching Policy (FM Policy)~\citep{zhang2024affordance} as multi-step controls; Consistency Policy (CP)~\citep{prasad2024consistency} and One-Step Diffusion Policy (OneDP)~\citep{wang2024one} representing distillation approaches; and MP1~\citep{sheng2025mp1} adapting MeanFlow~\citep{geng2025mean}. For 3D settings, we utilize DP3~\citep{ze20243d} and adapt FM Policy with an identical point encoder. CP and OneDP are distilled from a pre-trained DP3 teacher, whereas OFP and MP1 are trained from scratch. All evaluations share identical settings and are averaged over 3 random seeds (reported as mean $\pm$ standard deviation). Detailed experimental settings and evaluation metrics can be found in Appendices~\ref{app:policy_impl} and~\ref{app:simulation_experiments}.

\begin{table*}[t]
\centering
\caption{\textbf{Performance on 2D image-based manipulation.} Results are averaged over 3 random seeds. The best performance in each column is highlighted in \colorbox{yellow!25}{\textbf{bold}}.}
\label{tab:2d_results}
\resizebox{\textwidth}{!}{
\begin{tabular}{l | c | c c c c c c c | c}
\toprule
Method & NFE & Door & Hammer & Pen & Bucket & Faucet & Laptop & Toilet & Average \\
\midrule
DP & 100 & 51.7$\pm$4.0 & 100.0$\pm$0.0 & 67.7$\pm$1.2 & \cellcolor{yellow!25}\textbf{34.7$\pm$4.7} & 38.7$\pm$1.5 & 80.3$\pm$2.9 & 76.0$\pm$1.7 & 64.2$\pm$2.3 \\
DP & 10 & 44.7$\pm$1.5 & 100.0$\pm$0.0 & 66.7$\pm$1.5 & 32.7$\pm$4.0 & 35.0$\pm$2.6 & 81.0$\pm$2.6 & 66.0$\pm$3.6 & 60.9$\pm$2.3 \\
FM Policy & 100 & 63.7$\pm$2.1 & 100.0$\pm$0.0 & \cellcolor{yellow!25}\textbf{69.7$\pm$2.1} & 31.7$\pm$1.2 & 40.7$\pm$2.3 & 84.7$\pm$2.5 & 79.7$\pm$1.5 & 67.2$\pm$1.7 \\
FM Policy & 10 & 61.3$\pm$3.2 & 97.3$\pm$3.1 & 64.7$\pm$3.8 & 27.7$\pm$2.3 & 32.7$\pm$3.8 & \cellcolor{yellow!25}\textbf{87.0$\pm$1.7} & 77.3$\pm$1.5 & 64.0$\pm$2.8 \\
\midrule
CP & 1 & 46.3$\pm$3.5 & 100.0$\pm$0.0 & 58.7$\pm$5.7 & 30.7$\pm$2.5 & 33.3$\pm$0.6 & 79.0$\pm$2.6 & 69.7$\pm$3.1 & 59.7$\pm$2.6 \\
OneDP & 1 & 53.3$\pm$5.1 & 100.0$\pm$0.0 & 64.7$\pm$1.5 & 32.7$\pm$1.2 & 40.0$\pm$1.7 & 76.3$\pm$2.9 & 76.3$\pm$2.3 & 63.3$\pm$2.1 \\
MP1 & 1 & 49.3$\pm$6.0 & 96.3$\pm$3.5 & 57.3$\pm$4.9 & 31.7$\pm$6.7 & 33.0$\pm$7.5 & 83.7$\pm$1.5 & 72.3$\pm$4.6 & 60.5$\pm$5.0 \\
\textbf{OFP (Ours)} & 1 & \cellcolor{yellow!25}\textbf{68.3$\pm$3.1} & \cellcolor{yellow!25}\textbf{100.0$\pm$0.0} & 69.3$\pm$1.5 & 33.3$\pm$2.5 & \cellcolor{yellow!25}\textbf{41.0$\pm$1.7} & 85.7$\pm$3.5 & \cellcolor{yellow!25}\textbf{80.7$\pm$2.5} & \cellcolor{yellow!25}\textbf{68.3$\pm$2.1} \\
\bottomrule
\end{tabular}
}
\vspace{-5pt}
\end{table*}

\begin{table*}[t]
\centering
\caption{\textbf{Performance on 3D pointcloud manipulation.} We assess performance on 56 challenging tasks with 3 random seeds.}
\label{tab:3d_results}
\resizebox{\textwidth}{!}{
\begin{tabular}{l | c | c c c c c c | c}
\toprule
Method & NFE & \begin{tabular}{@{}c@{}}Adroit \\ (3)\end{tabular} & \begin{tabular}{@{}c@{}}DexArt \\ (4)\end{tabular} & \begin{tabular}{@{}c@{}}MetaWorld \\ Easy (28)\end{tabular} & \begin{tabular}{@{}c@{}}MetaWorld \\ Medium (11)\end{tabular} & \begin{tabular}{@{}c@{}}MetaWorld \\ Hard (5)\end{tabular} & \begin{tabular}{@{}c@{}}MetaWorld \\ Very Hard (5)\end{tabular} & Average \\
\midrule
DP3 & 100 & 79.0$\pm$0.6 & 63.7$\pm$0.6 & 82.8$\pm$8.0 & 46.5$\pm$3.8 & 38.3$\pm$9.6 & 41.0$\pm$0.6 & 66.4$\pm$5.7 \\
DP3 & 10 & 76.0$\pm$3.3 & 61.5$\pm$0.9 & 80.8$\pm$2.1 & 46.2$\pm$1.4 & 36.3$\pm$8.2 & 45.3$\pm$5.7 & 65.2$\pm$2.8 \\
FM Policy & 100 & 83.3$\pm$0.7 & 61.7$\pm$4.2 & 68.9$\pm$5.7 & 42.3$\pm$3.8 & \cellcolor{yellow!25}\textbf{43.6$\pm$3.4} & 47.7$\pm$4.5 & 59.8$\pm$4.6 \\
FM Policy & 10 & 80.3$\pm$1.8 & 60.5$\pm$4.5 & 66.7$\pm$7.7 & 41.3$\pm$7.4 & 40.3$\pm$7.2 & 46.7$\pm$7.6 & 57.9$\pm$7.0 \\
\midrule
CP & 1 & 76.3$\pm$4.6 & 60.7$\pm$3.4 & 68.3$\pm$1.8 & 32.1$\pm$4.3 & 28.7$\pm$5.2 & 36.0$\pm$2.3 & 54.7$\pm$2.9 \\
OneDP & 1 & 77.3$\pm$1.8 & 62.5$\pm$4.1 & 77.7$\pm$0.5 & 39.6$\pm$9.2 & 38.7$\pm$4.4 & 41.7$\pm$1.6 & 62.4$\pm$3.0 \\
MP1 & 1 & 82.0$\pm$2.1 & 61.0$\pm$2.0 & 70.5$\pm$6.1 & 37.9$\pm$5.9 & 34.3$\pm$9.5 & 32.3$\pm$5.5 & 57.4$\pm$5.8 \\
\textbf{OFP (Ours)} & 1 & \cellcolor{yellow!25}\textbf{85.0$\pm$4.5} & \cellcolor{yellow!25}\textbf{64.3$\pm$2.5} & \cellcolor{yellow!25}\textbf{87.9$\pm$4.6} & \cellcolor{yellow!25}\textbf{52.4$\pm$5.7} & 43.3$\pm$2.4 & \cellcolor{yellow!25}\textbf{49.2$\pm$0.7} & \cellcolor{yellow!25}\textbf{71.6$\pm$4.1} \\
\bottomrule
\end{tabular}
}
\vspace{-15pt}
\end{table*}

\vspace{-5pt}
\subsection{Experimental Results}

\textbf{2D Image-Conditioned Evaluation.} \cref{tab:2d_results} reports the results on the 2D tasks. OFP achieves the best average success rate among all one-step methods. It also surpasses the strong multi-step baselines. At NFE$=1$, OFP reaches $68.3\%$, which is higher than DP at NFE$=100$ with $64.2\%$ and FM Policy at NFE$=100$ with $67.2\%$. This shows that the proposed self-distillation objective does more than reduce latency. It also significantly improves final action quality. During training, MP1 showed high variance and frequent loss spikes. This is consistent with the extra numerical sensitivity introduced by the JVP operations in its objective. OFP avoids JVPs by training on an interval-averaged target, which leads to more stable optimization and better control accuracy.

\begin{table}[h]
\centering
\captionsetup{skip=4pt}
\footnotesize
\caption{\textbf{Few-Step vs. One-Step Generation.} OFP supports flexible sampling, maintaining high performance at 1 step while reliably improving accuracy given more generation steps.}
\label{tab:few_step}
\begin{tabular}{l | c | c c c c | c}
\toprule[0.6pt]
Method & NFE & Bucket & Faucet & Laptop & Toilet & Average \\
\midrule[0.3pt]
OneDP & 1 & 32.7$\pm$5.8 & 44.7$\pm$3.1 & 89.3$\pm$3.8 & \cellcolor{yellow!25}\textbf{83.3$\pm$3.8} & 62.5$\pm$4.1 \\
\midrule[0.3pt]
\multirow{2}{*}{CP} 
& 1 & 29.3$\pm$3.5 & 44.3$\pm$3.8 & 86.7$\pm$2.5 & 82.3$\pm$3.8 & 60.7$\pm$3.4 \\
& 4 & 38.3$\pm$4.2 & 43.7$\pm$1.2 & 90.0$\pm$3.5 & 84.7$\pm$2.9 & 64.2$\pm$2.9 \\
\midrule[0.3pt]
\multirow{2}{*}{OFP} 
& 1 & 39.3$\pm$0.6 & 45.7$\pm$3.8 & 91.7$\pm$3.2 & 81.3$\pm$3.2 & 64.5$\pm$2.7 \\
& 4 & \cellcolor{yellow!25}\textbf{42.7$\pm$2.1} & \cellcolor{yellow!25}\textbf{47.0$\pm$1.0} & \cellcolor{yellow!25}\textbf{92.3$\pm$4.2} & 83.0$\pm$1.7 & \cellcolor{yellow!25}\textbf{66.2$\pm$2.2} \\
\bottomrule[0.6pt]
\end{tabular}
\vspace{-10pt}
\end{table}

\textbf{3D Pointcloud-Conditioned Evaluation.} \cref{tab:3d_results} shows an even larger gain in the 3D setting. OFP achieves the highest success rates in the vast majority of tasks. At NFE$=1$, it outperforms DP3 at NFE$=100$ by $8\%$ on the average, and it exceeds the 3D FM Policy at NFE$=100$ by $19.7\%$. The gain is especially clear in the MetaWorld multi-task setting, where OFP improves both average performance and robustness across difficulty levels. These results support the role of self-guided regularization, which sharpens the action distribution and reduces the over-smoothing often seen in standard flow objectives.

\cref{fig:latency} shows the trade-off between inference time and success rate across all 56 tasks. OFP reaches the best average success rate while staying in the one-step low-latency regime. In wall-clock time, OFP requires $17.58$ ms per action chunk, compared with $3225.67$ ms for DP3 at NFE$=100$ and $1865.72$ ms for 3D FM Policy at NFE$=100$. This corresponds to about $183\times$ and $106\times$ speedups, respectively. OFP is slightly slower than OneDP and MP1 because warm-start adds a small amount of computation, but the accuracy gain is large enough to justify that cost.

\begin{wrapfigure}{r}{0.45\textwidth}
    \vspace{-18pt}
    \centering
    \includegraphics[width=\linewidth]{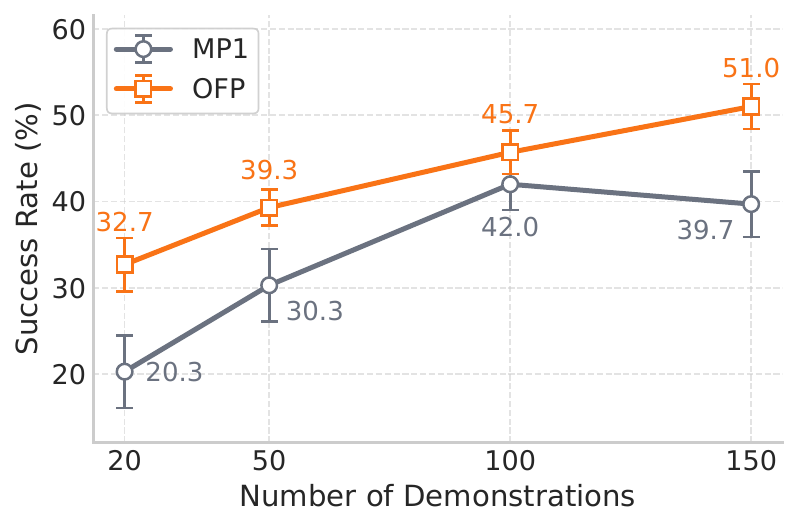} 
    \caption{\textbf{Data Scaling Behavior.} OFP extracts higher utility from sparse data (20 demos) and continues to scale cleanly as data increases, avoiding the performance degradation seen in MP1 at 150 demos.}
    \label{fig:data_scaling}
    \vspace{-16pt}
\end{wrapfigure}

\textbf{Latency and Accuracy Trade-off.} A basic limitation of OneDP-like methods is that they only support one-step inference. By contrast, trajectory-matching methods such as CP support multi-step generation but often suffer less precise single-step prediction. OFP combines the strengths of both. As shown in \cref{tab:few_step}, OneDP is stronger than CP at NFE$=1$ on several tasks. However, CP improves once more steps are allowed. OFP remains strong at NFE$=1$ and improves further at NFE$=4$, which shows that the same model can provide both low-latency one-step control and better accuracy when a slightly larger compute budget is available.

\textbf{Ablation Studies.} To isolate the contributions of our core components, we conducted targeted ablation studies. Specifically, we independently ablate each module under fixed training conditions to evaluate its impact on single-step and few-step success rates. Our results show that self-consistency training is necessary for reliable few-step inference, while self-guided regularization drives the performance gains in the single-step regime. Empirically, self-consistency and self-guided training complement each other. Additionally, the warm-start mechanism acts as a training-free inference prior that consistently improves generation quality across any step count. Unifying these mechanisms allows OFP to deliver high accuracy in both single-step and few-step control. We report the detailed results of these ablations in Appendix~\ref{app:ablation}.

\begin{figure}[t]
    \centering
    \includegraphics[width=\linewidth]{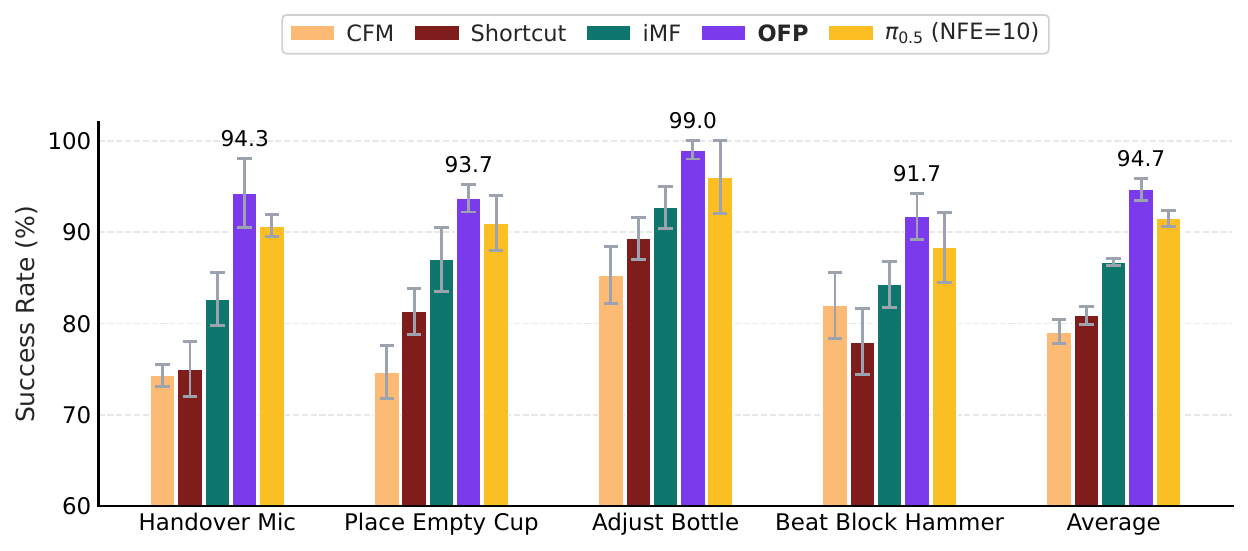} 
    \caption{\textbf{Transfer to $\pi_{0.5}$ on RoboTwin 2.0.} All acceleration methods are evaluated at NFE=1, and the $\pi_{0.5}$ baseline operates at NFE=10. OFP achieves the best average success rate across four tasks, showing that OFP remains effective even for large-scale VLA models with richer multi-modal inputs.}
    \label{fig:vla_integration}
    \vspace{-5pt}
\end{figure}

\begin{figure}[t]
    \centering
    \captionsetup{skip=4pt}
    \includegraphics[width=\linewidth]{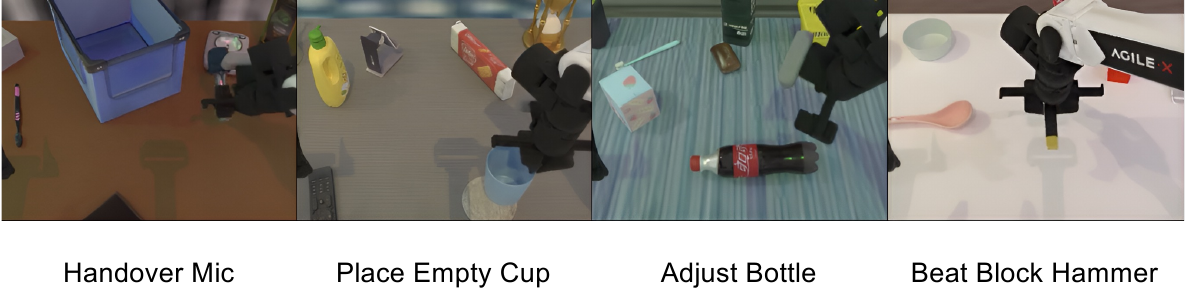} 
    \caption{\textbf{Visualization of RoboTwin 2.0 tasks.} Evaluations use domain randomization to test robustness under visual variation.}
    \label{fig:robotwin}
    \vspace{-15pt}
\end{figure}

\textbf{Data Efficiency and Scaling.} We investigate how performance scales with data availability on the DexArt Faucet task (\cref{fig:data_scaling}). Under very limited data, with only 20 demonstrations, MP1 degrades sharply. In contrast, OFP maintains a robust success rate, indicating that self-guided regularization provides stronger learning signals in the low-data regime. As the number of demonstrations increases, OFP improves steadily from $32.7\%$ to $51.0\%$. MP1 improves at first, but then drops after 100 demonstrations.

\subsection{Integration with VLA Models} 
\vspace{-6pt}
Modern VLA systems often require large over-parameterized models and condition action generation on rich multi-modal semantics. However, bootstrapping-based methods may suffer from rank-collapse when model capacity increases~\citep{frans2024one,kumar2020implicit}. To verify whether OFP remains effective in this regime, we integrate it into the $\pi_{0.5}$ model and evaluate it in RoboTwin 2.0. \cref{fig:robotwin} visualizes the 4 RoboTwin 2.0 tasks used in our $\pi_{0.5}$ integration study. We evaluate OFP under domain randomization (clutter/distractors, background textures, lighting, and tabletop height). We compare with three representative from-scratch acceleration methods: Consistency Flow Matching (CFM)~\citep{yang2024consistency}, Shortcut Models~\citep{frans2024one}, and Improved MeanFlow (iMF)~\citep{geng2025improved}.

\cref{fig:vla_integration} shows that OFP achieves the best average success rate, reaching $94.7\%$ across the four tasks. More importantly, OFP at \mbox{NFE$=1$} surpasses the original $\pi_{0.5}$ policy at \mbox{NFE$=10$}. This result shows that OFP does not collapse in a larger VLA setting with richer semantics. Instead, it preserves the benefits of single-step generation while still improving control quality. This confirms that OFP is not tied to a narrow policy class. The self-distillation design continues to work when the backbone is larger, the conditioning is richer, and the task semantics are more complex.

\section{Related Work}
\vspace{-3pt}
\subsection{Generative Policies for Robot Control}
The integration of diffusion models into robotics initially emerged through the planning-as-generation paradigm \citep{janner2022planning,ajay2022conditional}. Subsequently, the focus shifted toward visuomotor imitation: Diffusion Policy \citep{chi2025diffusion} established a robust baseline by parameterizing conditional policies as iterative denoising processes over action chunks. While this approach excels at modeling multimodal distributions and ensures training stability, it incurs high inference latency that scales linearly with the number of denoising steps. Subsequent research focuses on stronger conditional representations \citep{li2024crossway,ze20243d} and more structured generation \citep{ma2024hierarchical}. In parallel, Flow Matching policies \citep{hu2024adaflow,funk2024actionflow,zhang2025flowpolicy,braun2024riemannian,zhang2024affordance,chisari2024learning} leverage an ODE-based velocity field formulation. The flow-based paradigm is gradually converging with modern Vision-Language-Action (VLA) architectures \citep{liu2024rdt,black2024pi_0,shi2025hi,intelligence2025pi_,intelligence2025pi,bjorck2025gr00t}.

\vspace{-4pt}
\subsection{Distillation and Acceleration}
Few-step acceleration of generative policies mainly follows two routes.

\textbf{Consistency Distillation}. Building on Consistency Models \citep{song2023consistency}, approaches like Consistency Policy \citep{prasad2024consistency} and ManiCM \citep{lu2024manicm} distill a pre-trained diffusion teacher into a student. However, they are more likely to exhibit averaging under highly multimodal action distributions, resulting in insufficient one-step action precision.

\textbf{Score Distillation}. Inspired by Distribution Matching Distillation (DMD) \citep{yin2024one}, methods such as One-Step Diffusion Policy \citep{wang2024one} and SDM Policy \citep{jia2024score} utilize score-based gradients to optimize a one-step generator. While this approach produces sharper, mode-seeking samples ideal for precision tasks, it typically sacrifices distribution diversity and relies heavily on pre-trained teacher models.

Our approach unifies these two paradigms within a from-scratch self-distillation framework, achieving high-fidelity few-step and single-step generation without an external teacher. See Appendix~\ref{app:related_work} for a detailed version.

\vspace{-4pt}
\section{Conclusion \& Limitation}
We introduce OFP, a self-distillation framework that accelerates flow-based visuomotor policies to the one-step regime while maintaining high control precision. By unifying self-consistency and self-guided training, OFP achieves over a $100\times$ speedup against multi-step baselines, setting state-of-the-art success rates across diverse manipulation benchmarks and transferring effectively to larger VLA backbones like $\pi_{0.5}$. While OFP demonstrates robust performance across diverse tasks, our current evaluations are conducted in simulation. A direct next step is to evaluate OFP on physical robot systems. OFP is orthogonal to system-level acceleration techniques. Future work can seamlessly integrate OFP with model quantization and structural pruning to further decrease inference latency.

% \section*{References}
\bibliographystyle{unsrtnat}
\bibliography{references}

%%%%%%%%%%%%%%%%%%%%%%%%%%%%%%%%%%%%%%%%%%%%%%%%%%%%%%%%%%%%
\clearpage
\appendix

\section{Theoretical Supports and Proofs}
\subsection{Self-Consistency Training}
\label{app:proof_prop1}

We now provide a formal derivation for Proposition 1. The goal is to show that the target in self-consistency loss,
\[
\mathbf{u}_{\text{target}}
=
\frac{\hat{\mathbf z}_r-\mathbf z_t}{r-t},
\quad
\hat{\mathbf z}_r
=
\mathbf z_m+(r-m)\mathbf u_{\theta^-}(\mathbf z_m,m,r\mid o),
\]
is a consistent supervision signal for the ground-truth interval-velocity field
$\mathbf u(\mathbf z_t,t,r\mid o)$.

\paragraph{Definition of the ground-truth interval velocity.}
Let $\{\mathbf z_\tau\}_{\tau\in[0,1]}$ denote the true conditional transport trajectory associated with a fixed pair $(o,a,\epsilon)$. For any $0\le t<r\le 1$, define the interval-averaged velocity field
\begin{equation}
\mathbf u(\mathbf z_t,t,r\mid o)
\triangleq
\frac{\mathbf z_r-\mathbf z_t}{r-t}.
\label{eq:app_true_interval_velocity}
\end{equation}
By definition, it satisfies
\[
\mathbf z_r
=
\mathbf z_t+(r-t)\mathbf u(\mathbf z_t,t,r\mid o).
\]
Likewise, for any intermediate time $m\in[t,r]$ on the same trajectory,
\[
\mathbf z_r
=
\mathbf z_m+(r-m)\mathbf u(\mathbf z_m,m,r\mid o).
\]

\paragraph{Assumptions.}
We work under the following assumptions.

\begin{enumerate}
    \item[(A1)] The ground-truth interval-velocity field
    $\mathbf u(\mathbf x,t,r\mid o)$ is continuously differentiable in $(\mathbf x,t,r)$ on the domain
    $\{(\mathbf x,t,r):0\le t\le r\le 1\}$.

    \item[(A2)] $\mathbf u$ is globally Lipschitz continuous in state:
    there exists $L_x>0$ such that for all $\mathbf x,\mathbf y$,
    \[
    \|\mathbf u(\mathbf x,t,r\mid o)-\mathbf u(\mathbf y,t,r\mid o)\|
    \le L_x\|\mathbf x-\mathbf y\|.
    \]

    \item[(A3)] The EMA teacher is accurate. Specifically, define
    \[
    \varepsilon_{\mathrm{ema}}(s)
    \triangleq
    \sup_{o,\mathbf x,t,r}
    \|\mathbf u_{\theta^-}(\mathbf x,t,r\mid o)-\mathbf u(\mathbf x,t,r\mid o)\|,
    \]
    and assume $\varepsilon_{\mathrm{ema}}(s)\to 0$ as training proceeds.

    \item[(A4)] The intermediate time $m$ is sampled from
    \[
    m\sim \mathcal U\!\Big[t,\; t+(r-t)\rho(s)\Big],
    \]
    where $\rho(s)\to 0$ as the training step $s\to\infty$.
\end{enumerate}

\paragraph{Algebraic decomposition of the target.}
Starting from the definition of $\mathbf u_{\text{target}}$,
\[
\mathbf u_{\text{target}}
=
\frac{\mathbf z_m-\mathbf z_t}{r-t}
+
\frac{r-m}{r-t}\mathbf u_{\theta^-}(\mathbf z_m,m,r\mid o).
\]
On the other hand, using the exact interval update from $m$ to $r$,
\[
\mathbf z_r-\mathbf z_t
=
(\mathbf z_m-\mathbf z_t)
+
(r-m)\mathbf u(\mathbf z_m,m,r\mid o).
\]
Dividing both sides by $(r-t)$ gives
\[
\mathbf u(\mathbf z_t,t,r\mid o)
=
\frac{\mathbf z_m-\mathbf z_t}{r-t}
+
\frac{r-m}{r-t}\mathbf u(\mathbf z_m,m,r\mid o).
\]
Subtracting the two expressions yields
\begin{equation}
\mathbf u_{\text{target}}-\mathbf u(\mathbf z_t,t,r\mid o)
=
\frac{r-m}{r-t}
\Big(
\mathbf u_{\theta^-}(\mathbf z_m,m,r\mid o)
-
\mathbf u(\mathbf z_m,m,r\mid o)
\Big).
\label{eq:app_main_identity}
\end{equation}

\paragraph{Pointwise consistency of the target.}
Since $m\in[t,r]$, we have $0\le (r-m)/(r-t)\le 1$. Therefore, from \cref{eq:app_main_identity},
\[
\bigl\|
\mathbf u_{\text{target}}-\mathbf u(\mathbf z_t,t,r\mid o)
\bigr\|
\le
\bigl\|
\mathbf u_{\theta^-}(\mathbf z_m,m,r\mid o)
-
\mathbf u(\mathbf z_m,m,r\mid o)
\bigr\|
\le
\varepsilon_{\mathrm{ema}}(s).
\]
Hence,
\begin{equation}
\mathbf u_{\text{target}}
\;\xrightarrow[s\to\infty]{}\;
\mathbf u(\mathbf z_t,t,r\mid o)
\quad
\text{uniformly on the training domain.}
\label{eq:app_target_consistency}
\end{equation}

This proves that the teacher-generated target is a consistent supervision signal whenever the EMA teacher is accurate.

\paragraph{Role of the time-contracting schedule.}
Since $m\in[t,t+(r-t)\rho(s)]$, we have $0\le m-t\le (r-t)\rho(s).$
Because the true trajectory is continuously differentiable, there exists a constant $C_z>0$ such that
\[
\|\mathbf z_m-\mathbf z_t\|
\le
C_z |m-t|
\le
C_z (r-t)\rho(s).
\]
Using continuous differentiability in $(t,r)$ and Lipschitz continuity in state, there exists a constant $C_u>0$ such that
\[
\|\mathbf u(\mathbf z_m,m,r\mid o)-\mathbf u(\mathbf z_t,t,r\mid o)\|
\le
C_u\bigl(\|\mathbf z_m-\mathbf z_t\|+|m-t|\bigr)
\le
C_u'(r-t)\rho(s),
\]
for some constant $C_u'>0$.

Thus, as $\rho(s)\to 0$, the teacher query point $(\mathbf z_m,m,r)$ approaches $(\mathbf z_t,t,r)$, and $\mathbf u(\mathbf z_m,m,r\mid o)\to \mathbf u(\mathbf z_t,t,r\mid o).$
This shows that the contracting schedule gradually turns the target from a coarse long-interval bootstrap signal into a local self-consistency signal centered at the student input $(\mathbf z_t,t,r)$.

\paragraph{Convergence of the population regression target.}
Let $X\triangleq (\mathbf z_t,t,r,o), \ Y\triangleq \mathbf u_{\text{target}}.$
Write the target as
\[
Y=\mathbf u(X)+\boldsymbol{\eta},
\quad
\boldsymbol{\eta}
\triangleq
\mathbf u_{\text{target}}-\mathbf u(X).
\]
By \cref{eq:app_main_identity}, $\|\boldsymbol{\eta}\|\le \varepsilon_{\mathrm{ema}}(s)$, and thus
\[
\mathbb E\|\boldsymbol{\eta}\|^2 \le \varepsilon_{\mathrm{ema}}(s)^2 \to 0.
\]

Consider the population self-consistency loss
\[
\mathcal L_{\text{self-consistency}}(\theta)
=
\mathbb E\bigl[\|\mathbf u_\theta(X)-Y\|^2\bigr].
\]
Let $\theta^\star$ be any global minimizer of this objective. Since $\mathbf u$ is a valid comparator,
\[
\mathcal L_{\text{self-consistency}}(\theta^\star)
\le
\mathcal L_{\text{self-consistency}}(\mathbf u)
=
\mathbb E\|\boldsymbol{\eta}\|^2.
\]
Expanding the loss at $\theta^\star$ gives
\[
\mathcal L_{\text{self-consistency}}(\theta^\star)
=
\mathbb E\|\mathbf u_{\theta^\star}(X)-\mathbf u(X)\|^2
+
2\mathbb E\bigl[\langle \mathbf u_{\theta^\star}(X)-\mathbf u(X),\boldsymbol{\eta}\rangle\bigr]
+
\mathbb E\|\boldsymbol{\eta}\|^2.
\]
Combining the two displays and applying Cauchy--Schwarz,
\[
\mathbb E\|\mathbf u_{\theta^\star}(X)-\mathbf u(X)\|^2
\le
2
\Big(
\mathbb E\|\mathbf u_{\theta^\star}(X)-\mathbf u(X)\|^2
\Big)^{1/2}
\Big(
\mathbb E\|\boldsymbol{\eta}\|^2
\Big)^{1/2}.
\]
Therefore,
\[
\Big(
\mathbb E\|\mathbf u_{\theta^\star}(X)-\mathbf u(X)\|^2
\Big)^{1/2}
\le
2
\Big(
\mathbb E\|\boldsymbol{\eta}\|^2
\Big)^{1/2},
\]
which implies
\begin{equation}
\mathbb E\|\mathbf u_{\theta^\star}(X)-\mathbf u(X)\|^2
\le
4\,\mathbb E\|\boldsymbol{\eta}\|^2
\;\xrightarrow[s\to\infty]{}\;0.
\label{eq:app_l2_convergence}
\end{equation}
Hence,
\begin{equation}
\mathbf u_{\theta^\star}
\to
\mathbf u
\quad
\text{in }L^2.
\label{eq:app_final_conclusion}
\end{equation}

\paragraph{Conclusion.}
Equations \cref{eq:app_main_identity,eq:app_target_consistency,eq:app_l2_convergence,eq:app_final_conclusion} prove the claim in Proposition 1: when the EMA teacher is accurate and the contracting factor satisfies $\rho(s)\to 0$, the target $\mathbf u_{\text{target}}$ is a consistent supervision signal for the true interval-velocity field $\mathbf u(\mathbf z_t,t,r\mid o)$, and minimizing the self-consistency loss recovers $\mathbf u_\theta\approx \mathbf u$.

% \hfill $\square$

\subsection{Relation to MeanFlow}
\label{app:meanflow_equivalence}

In this section, we show that our consistency target admits a first-order connection to the differential term used in MeanFlow\cite{geng2025mean}. In particular, as the sub-interval length $(m-t)\to 0$, our target induces a finite-difference approximation of the total derivative appearing in the MeanFlow identity. Unlike MeanFlow, which computes this derivative through a Jacobian-vector product (JVP), our construction uses only forward evaluations of the EMA teacher. MeanFlow explicitly derives the average-velocity identity and then expands its total derivative as a JVP term during training.

\paragraph{Setup.}
Fix a conditional sample $(o,a)$ and noise $\epsilon$. Along the straight OT conditional path, $\mathbf z_\tau = (1-\tau)\epsilon + \tau a, \ \tau\in[0,1],$
the instantaneous velocity is constant:
$
\dot{\mathbf z}_\tau = a-\epsilon \triangleq \mathbf v.
$
Define the ground-truth interval-averaged velocity field by
\[
\mathbf u(\mathbf z_t,t,r\mid o)
\triangleq
\frac{\mathbf z_r-\mathbf z_t}{r-t},
\quad 0\le t<r\le 1.
\]

\paragraph{MeanFlow identity.}
Starting from
\[
(r-t)\,\mathbf u(\mathbf z_t,t,r\mid o)=\mathbf z_r-\mathbf z_t,
\]
differentiate both sides with respect to $t$ along the trajectory, while holding $r$ fixed. Since $\frac{d}{dt}\mathbf z_t=\mathbf v$, the product rule gives
\[
-\mathbf u(\mathbf z_t,t,r\mid o)
+
(r-t)\frac{d}{dt}\mathbf u(\mathbf z_t,t,r\mid o)
=
-\mathbf v.
\]
Rearranging yields
\begin{equation}
\mathbf u(\mathbf z_t,t,r\mid o)
=
\mathbf v
+
(r-t)\frac{d}{dt}\mathbf u(\mathbf z_t,t,r\mid o).
\label{eq:app_meanflow_identity}
\end{equation}
Equivalently,
\begin{equation}
\frac{d}{dt}\mathbf u(\mathbf z_t,t,r\mid o)
=
\frac{\mathbf u(\mathbf z_t,t,r\mid o)-\mathbf v}{r-t}.
\label{eq:app_meanflow_derivative}
\end{equation}
This is exactly the differential relation used by MeanFlow. In MeanFlow, the total derivative is further expanded into a partial-time term plus a state Jacobian term, which is then implemented by a JVP.

\paragraph{Consistency target with an exact teacher.}
For the purpose of analysis, first replace the EMA teacher with the exact field. Then our consistency target becomes
\[
\mathbf u_{\text{target}}^\star
=
\frac{\mathbf z_m+(r-m)\mathbf u(\mathbf z_m,m,r\mid o)-\mathbf z_t}{r-t},
\quad t<m<r.
\]
Because the OT path is linear, $\mathbf z_m-\mathbf z_t=(m-t)\mathbf v.$
Substituting this into the target gives
\[
\mathbf u_{\text{target}}^\star
=
\frac{m-t}{r-t}\mathbf v
+
\frac{r-m}{r-t}\mathbf u(\mathbf z_m,m,r\mid o).
\]
Subtract \(\mathbf u(\mathbf z_m,m,r\mid o)\) from both sides:
\[
\mathbf u_{\text{target}}^\star-\mathbf u(\mathbf z_m,m,r\mid o)
=
\frac{m-t}{r-t}
\Big(
\mathbf v-\mathbf u(\mathbf z_m,m,r\mid o)
\Big).
\]
Hence,
\begin{equation}
\frac{\mathbf u(\mathbf z_m,m,r\mid o)-\mathbf u_{\text{target}}^\star}{m-t}
=
\frac{\mathbf u(\mathbf z_m,m,r\mid o)-\mathbf v}{r-t}.
\label{eq:app_fd_exact}
\end{equation}
Applying the MeanFlow derivative identity at time \(m\),
\[
\frac{d}{dm}\mathbf u(\mathbf z_m,m,r\mid o)
=
\frac{\mathbf u(\mathbf z_m,m,r\mid o)-\mathbf v}{r-m},
\]
we can rewrite the right-hand side of \eqref{eq:app_fd_exact} as
\[
\frac{\mathbf u(\mathbf z_m,m,r\mid o)-\mathbf v}{r-t}
=
\frac{r-m}{r-t}\,
\frac{d}{dm}\mathbf u(\mathbf z_m,m,r\mid o).
\]
Therefore,
\begin{equation}
\frac{\mathbf u(\mathbf z_m,m,r\mid o)-\mathbf u_{\text{target}}^\star}{m-t}
=
\frac{r-m}{r-t}\,
\frac{d}{dm}\mathbf u(\mathbf z_m,m,r\mid o).
\label{eq:app_fd_to_derivative}
\end{equation}

\paragraph{Limit as \(m\to t^+\).}
Assume \(\mathbf u\) is continuously differentiable. Taking the limit \(m\to t^+\) in \eqref{eq:app_fd_to_derivative} yields
\begin{equation}
\lim_{m\to t^+}
\frac{\mathbf u(\mathbf z_m,m,r\mid o)-\mathbf u_{\text{target}}^\star}{m-t}
=
\frac{d}{dt}\mathbf u(\mathbf z_t,t,r\mid o).
\label{eq:app_limit_derivative}
\end{equation}
Thus, the discrepancy between the exact consistency target and the field value at the later point \((\mathbf z_m,m,r)\) is, after division by the small interval length \((m-t)\), a finite-difference approximation of the MeanFlow derivative term.

\paragraph{Including the EMA teacher.}
Now return to the practical target
\[
\mathbf u_{\text{target}}
=
\frac{\mathbf z_m+(r-m)\mathbf u_{\theta^-}(\mathbf z_m,m,r\mid o)-\mathbf z_t}{r-t}.
\]
Let
\[
\boldsymbol\delta_m
\triangleq
\mathbf u_{\theta^-}(\mathbf z_m,m,r\mid o)-\mathbf u(\mathbf z_m,m,r\mid o).
\]
Then
\[
\mathbf u_{\text{target}}
=
\mathbf u_{\text{target}}^\star
+
\frac{r-m}{r-t}\boldsymbol\delta_m.
\]
Therefore,
\[
\frac{\mathbf u(\mathbf z_m,m,r\mid o)-\mathbf u_{\text{target}}}{m-t}
=
\frac{\mathbf u(\mathbf z_m,m,r\mid o)-\mathbf u_{\text{target}}^\star}{m-t}
-
\frac{r-m}{r-t}\frac{\boldsymbol\delta_m}{m-t}.
\]
If the EMA teacher is accurate and its approximation error satisfies
\[
\|\boldsymbol\delta_m\| = o(m-t)
\quad \text{as } m\to t^+,
\]
then the second term vanishes, and \eqref{eq:app_limit_derivative} still holds with \(\mathbf u_{\text{target}}\) in place of \(\mathbf u_{\text{target}}^\star\). Consequently, in the small-interval limit, our consistency target recovers the same first-order differential signal that MeanFlow computes through a JVP, but it does so using only forward teacher evaluations.

\paragraph{Conclusion.}
Equation \eqref{eq:app_limit_derivative} shows the precise relation: as \((m-t)\to 0\), our consistency target induces a finite-difference approximation to the total derivative term in the MeanFlow identity. Hence, our method can be viewed as a forward-only discretization of the MeanFlow differential objective. This explains why the two formulations become equivalent in the local limit, while our implementation avoids the explicit JVP used in MeanFlow training. MeanFlow’s training target is built from the average-velocity identity and computes the derivative term through JVP.

% \hfill $\square$

\subsection{Gradient Alignment of Self-Guidance with the CA Term}
\label{app:self_guidance_ca}

In this section, we prove that minimizing the self-guidance loss yields a gradient whose descent direction is aligned with the CFG Augmentation (CA) term. The result formalizes the claim that the self-guidance objective acts as a self-distilled version of the CA mechanism used in score-distillation methods.

\paragraph{Notation.}
For brevity, define the one-step predictor
\[
\mathbf f_\theta
\triangleq
\mathbf u_\theta(\mathbf z_t,t,1\mid o),
\quad
\hat{\mathbf a}
=
\mathbf z_t+(1-t)\mathbf f_\theta .
\]
The re-noised sample is $
\tilde{\mathbf z}_{t'}
=
(1-t')\epsilon_1+t'\hat{\mathbf a},
\
t'\sim\mathcal U[0,1]. $ Let
\[
\mathbf u^-_{\mathrm{cond}}
\triangleq
\mathbf u_{\theta^-}(\tilde{\mathbf z}_{t'},t',t'\mid o),
\quad
\mathbf u^-_{\mathrm{uncond}}
\triangleq
\mathbf u_{\theta^-}(\tilde{\mathbf z}_{t'},t',t'\mid \phi).
\]
Then the target used in self-guidance is
\[
\mathbf s_{\mathrm{target}}
=
\mathrm{sg}\!\left[
\mathbf f_\theta
-
\Big(
\mathbf u^-_{\mathrm{uncond}}
-
\mathbf u^-_{\mathrm{cond}}
\Big)
\right],
\]
where $\mathrm{sg}[\cdot]$ denotes stop-gradient.

\paragraph{Rewrite the loss in residual form.}
The self-guidance loss is
\[
\mathcal L_{\mathrm{self\text{-}guidance}}(\theta)
=
\mathbb E\Big[
\|\mathbf f_\theta-\mathbf s_{\mathrm{target}}\|_2^2
\Big].
\]
Define
\[
\boldsymbol\Delta_{\mathrm{CA}}
\triangleq
\mathbf u^-_{\mathrm{uncond}}-\mathbf u^-_{\mathrm{cond}}.
\]
Since $\mathrm{sg}[\cdot]$ freezes its argument in backpropagation, the forward residual is
\[
\mathbf f_\theta-\mathbf s_{\mathrm{target}}
=
\mathbf f_\theta-\mathrm{sg}\!\left[\mathbf f_\theta-\boldsymbol\Delta_{\mathrm{CA}}\right].
\]
Numerically, this residual equals $\boldsymbol\Delta_{\mathrm{CA}}$, while its gradient flows only through the first \(\mathbf f_\theta\). Therefore,
\[
\mathcal L_{\mathrm{self\text{-}guidance}}(\theta)
=
\mathbb E\Big[
\|\mathbf f_\theta-\mathrm{sg}[\mathbf f_\theta-\boldsymbol\Delta_{\mathrm{CA}}]\|_2^2
\Big].
\]

\paragraph{Compute the gradient of the self-guidance loss.}
Let
\[
J_\theta(\mathbf z_t,t,o)
\triangleq
\frac{\partial \mathbf f_\theta}{\partial \theta}
\]
denote the Jacobian of the one-step predictor with respect to the model parameters. Since the stop-gradient branch contributes no derivative, the chain rule gives
\begin{equation}
\nabla_\theta \mathcal L_{\mathrm{self\text{-}guidance}}
=
2\,\mathbb E\!\left[
J_\theta(\mathbf z_t,t,o)^\top
\boldsymbol\Delta_{\mathrm{CA}}
\right].
\label{eq:app_self_guidance_grad}
\end{equation}
Hence, the gradient-descent update induced by the loss is
\begin{equation}
-\nabla_\theta \mathcal L_{\mathrm{self\text{-}guidance}}
=
-2\,\mathbb E\!\left[
J_\theta(\mathbf z_t,t,o)^\top
\boldsymbol\Delta_{\mathrm{CA}}
\right].
\label{eq:app_self_guidance_descent}
\end{equation}
This already shows that the update direction is directly driven by the EMA-based conditional--unconditional gap.

\paragraph{Express the CA term in the same variables.}
Under the OT path $\tilde{\mathbf z}_{t'}=(1-t')\epsilon+t'\mathbf a,$
the marginal score and instantaneous velocity satisfy
\[
\mathbf s(\tilde{\mathbf z}_{t'}\mid c)
=
\frac{t'\,\mathbf u(\tilde{\mathbf z}_{t'},t',t'\mid c)-\tilde{\mathbf z}_{t'}}{1-t'},
\]
for condition \(c\in\{o,\phi\}\). Therefore,
\[
\mathbf s(\tilde{\mathbf z}_{t'}\mid \phi)-\mathbf s(\tilde{\mathbf z}_{t'}\mid o)
=
\frac{t'}{1-t'}
\Big(
\mathbf u(\tilde{\mathbf z}_{t'},t',t'\mid \phi)
-
\mathbf u(\tilde{\mathbf z}_{t'},t',t'\mid o)
\Big).
\]
Using the EMA teacher to estimate the score, define the teacher-side CA vector
\begin{equation}
\mathbf g_{\mathrm{CA}}
\triangleq
\mathbf s^-_{\mathrm{uncond}}-\mathbf s^-_{\mathrm{cond}}
=
\frac{t'}{1-t'}\,
\boldsymbol\Delta_{\mathrm{CA}},
\label{eq:app_ca_score_velocity_relation}
\end{equation}
where
$
\mathbf s^-_{\mathrm{cond}}
\triangleq
\mathbf s_{\theta^-}(\tilde{\mathbf z}_{t'}\mid o),
\quad
\mathbf s^-_{\mathrm{uncond}}
\triangleq
\mathbf s_{\theta^-}(\tilde{\mathbf z}_{t'}\mid \phi).
$
Thus,
\begin{equation}
\boldsymbol\Delta_{\mathrm{CA}}
=
\frac{1-t'}{t'}\,\mathbf g_{\mathrm{CA}}.
\label{eq:app_velocity_from_ca}
\end{equation}

\paragraph{Compare with the CA gradient in score-distillation.}
In a reverse-KL-style score-distillation objective, the practical CFG-based gradient contains a CA term of the form
\[
\nabla_\theta \mathcal J_{\mathrm{CA}}
=
\mathbb E\!\left[
\mathbf g_{\mathrm{CA}}^\top
\frac{\partial \tilde{\mathbf z}_{t'}}{\partial \theta}
\right].
\]
Since
$
\tilde{\mathbf z}_{t'}
=
(1-t')\epsilon_1+t'\hat{\mathbf a},
\quad
\hat{\mathbf a}
=
\mathbf z_t+(1-t)\mathbf f_\theta,
$
we have
\[
\frac{\partial \tilde{\mathbf z}_{t'}}{\partial \theta}
=
t'(1-t)\,
J_\theta(\mathbf z_t,t,o).
\]
Substituting this together with \eqref{eq:app_velocity_from_ca} gives
\begin{align}
\nabla_\theta \mathcal J_{\mathrm{CA}}
&=
t'(1-t)\,
\mathbb E\!\left[
J_\theta(\mathbf z_t,t,o)^\top
\mathbf g_{\mathrm{CA}}
\right] \nonumber\\
&=
t'(1-t)\,
\mathbb E\!\left[
J_\theta(\mathbf z_t,t,o)^\top
\frac{t'}{1-t'}\boldsymbol\Delta_{\mathrm{CA}}
\right] \nonumber\\
&=
\frac{t'^2(1-t)}{1-t'}\,
\mathbb E\!\left[
J_\theta(\mathbf z_t,t,o)^\top
\boldsymbol\Delta_{\mathrm{CA}}
\right].
\label{eq:app_ca_grad_form}
\end{align}

Comparing \eqref{eq:app_self_guidance_grad} and \eqref{eq:app_ca_grad_form}, we obtain
\begin{equation}
\nabla_\theta \mathcal J_{\mathrm{CA}}
=
\frac{t'^2(1-t)}{2(1-t')}
\,
\nabla_\theta \mathcal L_{\mathrm{self\text{-}guidance}}.
\label{eq:app_gradient_proportionality}
\end{equation}
Therefore, the two gradients are proportional by a strictly positive scalar whenever \(t\in[0,1)\) and \(t'\in(0,1)\). Equivalently, their gradient-descent directions are identical:
\begin{equation}
-\nabla_\theta \mathcal L_{\mathrm{self\text{-}guidance}}
\;\parallel\;
-\nabla_\theta \mathcal J_{\mathrm{CA}}.
\label{eq:app_same_descent_direction}
\end{equation}

\paragraph{Interpretation.}
Equation \eqref{eq:app_same_descent_direction} proves that minimizing the self-guidance loss applies the same first-order update direction as the CA term, up to a positive rescaling. Since
\[
\boldsymbol\Delta_{\mathrm{CA}}
=
\mathbf u^-_{\mathrm{uncond}}-\mathbf u^-_{\mathrm{cond}},
\]
the descent update pushes the conditional one-step predictor away from the unconditional branch and toward the condition-specific direction favored by CFG.

\paragraph{Conclusion.}
The self-guidance loss yields a parameter gradient of the form
\[
\nabla_\theta \mathcal L_{\mathrm{self\text{-}guidance}}
=
2\,\mathbb E[J_\theta^\top\boldsymbol\Delta_{\mathrm{CA}}],
\]
while the CA term in score-distillation gives
\[
\nabla_\theta \mathcal J_{\mathrm{CA}}
\propto
\mathbb E[J_\theta^\top\boldsymbol\Delta_{\mathrm{CA}}].
\]
Hence, the two are equivalent up to a positive scalar factor. This proves that the self-guidance loss implements the CA effect in a self-distilled, forward-trainable form.

% \hfill $\square$

\section{Policy Implementation Details}
\label{app:policy_impl}

We detail the observation pipelines, network architectures, and optimization hyperparameters for evaluating OFP across the Adroit~\cite{kumar2016manipulators}, DexArt~\cite{bao2023dexart}, MetaWorld~\cite{yu2020meta}, and RoboTwin 2.0~\cite{chen2025robotwin} benchmarks.

\subsection{Observation Inputs}

\textbf{2D setting.}
Image-based control relies on a two-frame history of 84x84 RGB observations, a low-dimensional proprioceptive state (e.g., 24 dimensions for Adroit Hammer, 32 for DexArt Faucet), and an 8-step action chunk target. Receding-horizon control executes the first 4 actions. Visual features are extracted using a ResNet-18 backbone, trained end-to-end with the policy. Inputs undergo standard normalization and data augmentation (random crop, rotation, color jitter). The encoded visual and state features are concatenated into a global condition vector for the flow model.

\textbf{3D setting.}
Point-cloud control similarly uses a 2-step observation history and an 8-step action horizon. Input sizes range from 512 points (Adroit, MetaWorld) to 1024 scene points plus 96 imagined-robot points (DexArt). We process 3D observations using the DP3 encoder~\cite{ze20243d}, which employs a PointNet-based extractor and an MLP state encoder. The resulting 128-dimensional point-cloud features and 64-dimensional state features are concatenated to produce a short conditioning sequence for the transformer policy.

\textbf{$\pi_{0.5}+$OFP Integration.}
This setup utilizes three 224$\times$224 RGB views, a 32-dimensional robot state, and a PaliGemma-tokenized language prompt. Images are scaled to $[-1, 1]$ and augmented with random crops, resizes, rotations, and color jitter. States and actions are padded to match the model dimension.

\subsection{Model Architecture}

\textbf{Backbone.}
OFP employs a Diffusion Transformer (DiT)~\cite{peebles2023scalable}  to predict an interval-averaged velocity field, conditioned on the current action, start time, target interval, and observation embedding. Both 2D and 3D variants use 12 transformer layers, 8 attention heads, and a hidden dimension of 768. Timestep and target-interval embeddings (128 dimensions each) are concatenated and projected. Observations are integrated via cross-attention, and a linear head outputs the final action.

\textbf{Diffusion baselines.}
DP and the distilled baselines (CP, OneDP) use a U-Net denoiser with channel widths of $[256, 512, 1024]$. DP3 uses wider channels of $[512, 1024, 2048]$. All distilled baselines retain their respective teacher architectures.

\textbf{$\pi_{0.5}$+OFP architecture.}
The integration uses a two-expert PaliGemma architecture: a 2B-parameter vision-language prefix expert (incorporating a SigLIP encoder) and a 300M-parameter action expert. The prefix expert fuses multi-view cameras and language prompts into a shared token sequence. The action expert (depth 18, width 1024, 8 heads) maps noisy action tokens to flow updates. Time and interval conditioning are applied via adaptive RMS normalization.

\subsection{Training Details}

\textbf{OFP training.}
We optimize OFP using AdamW (learning rate 1e-4, beta1=0.95, beta2=0.999, weight decay 1e-6) with a cosine schedule and 500 warmup steps. The EMA teacher uses an exponential decay of 0.75 (max 0.9999). Training runs for 3000 epochs on a single NVIDIA A100 GPU with a batch size of 64. The training objective allocates 80\% weight to boundary flow-matching and 20\% to self-consistency. Self-consistency intervals are sampled from a logit-normal distribution, and the time-contraction schedule follows a polynomial decay. For 3D tasks, self-guidance is enabled with a weight of 0.05 and a null-condition dropout of 10\%. During inference, warm-start initialization is applied at $t_w=0.15, \ z_{t_w}=(1-t_w)\epsilon + t_w a_{\mathrm{warm}}.
$

\textbf{$\pi_{0.5}$+OFP training.}
Following the OpenPI~\cite{black2024pi_0} configuration, we use the AdamW optimizer with global gradient clipping at 1.0 and a cosine schedule peaking at a learning rate of 2.5e-5. The model is fine-tuned for 3000 steps from the base $\pi_{0.5}$ checkpoint using 4 NVIDIA A100 GPUs.

\section{Simulation Experiments}
\label{app:simulation_experiments}

\begin{figure}[h]
    \centering
    \includegraphics[width=\linewidth]{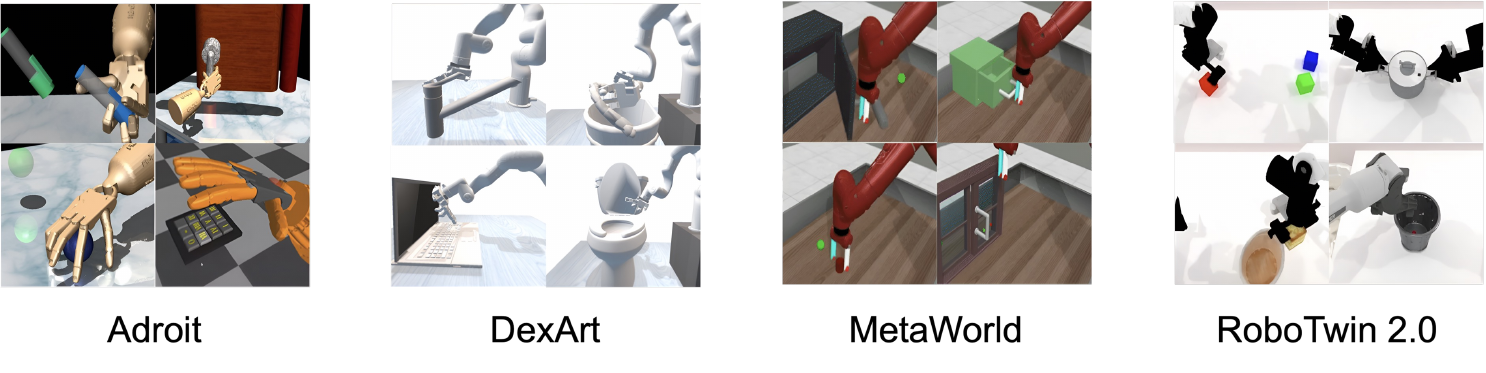} 
    \caption{\textbf{Visualizations of the simulated manipulation benchmarks.} These diverse environments rigorously test the capability of OFP across different robotic setups, geometric constraints, and visual conditions.}
    \label{fig:benchmarks}
\end{figure}

\subsection{Benchmarks}
We conduct evaluations on four distinct simulated environments, visualized in \cref{fig:benchmarks}.

\textbf{Adroit.}
Adroit~\cite{kumar2016manipulators} is a dexterous manipulation benchmark built on a high-degree-of-freedom anthropomorphic hand. It includes tasks such as opening a door with a latch, hammering a nail, and reorienting a pen. These tasks require precise finger coordination and accurate contact-rich control.

\textbf{DexArt.}
DexArt~\cite{bao2023dexart} is a benchmark for dexterous manipulation of articulated objects. It focuses on manipulation problems such as turning faucets, opening laptop lids, lifting buckets, and opening toilet lids. Compared with rigid-object tasks, articulated objects introduce additional motion constraints and stronger geometric variation.

\textbf{MetaWorld.}
MetaWorld~\cite{yu2020meta} is a large-scale simulated manipulation suite designed for multi-task and meta-learning. It contains a broad set of tabletop manipulation problems with diverse task semantics. We use a 49-task multi-task setup and group the tasks into four empirical difficulty levels: Easy, Medium, Hard, and Very Hard. A single policy is trained jointly across all 49 tasks.

\textbf{RoboTwin 2.0.}
RoboTwin 2.0~\cite{chen2025robotwin} is a dual-arm manipulation benchmark for robust bimanual control. A key property of RoboTwin 2.0 is strong domain randomization, including cluttered distractors, background textures, lighting conditions, and tabletop height changes. We use RoboTwin 2.0 to evaluate whether OFP can be integrated into a large VLA model and remain stable under semantically rich and visually varying bimanual tasks.

\subsection{Experimental Protocol}

\textbf{Adroit and DexArt.}
For both benchmarks, we train all methods for 3000 epochs. We use 10 demonstrations per task for Adroit, and 100 for DexArt. During training, we evaluate 20 rollout episodes every 50 epochs. Following the DP3~\cite{ze20243d} protocol, the final success rate is defined as the mean of the top-5 rollout success rates observed over training.

\textbf{MetaWorld.}
For the 49-task multi-task setting, we train all methods for 1000 epochs. We use 10 demonstrations per task and evaluate 20 episodes every 200 epochs. We define the final success rate as the mean of the top-3 rollout success rates within the training window.

\textbf{RoboTwin 2.0.}
For the $\pi_{0.5}$ integration study, we fine-tune the model for 3k training steps with 100 demonstrations per task. After training, we evaluate each policy on 100 randomized test trials.

\subsection{Full Results}

The detailed 3D results are reported in \cref{tab:app_sim_3d_1}. OFP consistently improves over the single-step baselines and remains competitive even against strong multi-step policies. These results show that OFP preserves the speed advantage of one-step generation while maintaining high action quality on control problems.

The full $\pi_{0.5}$ integration results are shown in \cref{tab:app_robotwin_full}. OFP achieves the best performance on all 4 RoboTwin 2.0 tasks and reaches the highest average success rate. More importantly, OFP at NFE$=1$ outperforms the original $\pi_{0.5}$ policy at NFE$=10$ on every task. This shows that the proposed self-distillation design remains effective after being inserted into a large VLA backbone and under strong domain randomization.

\begin{table*}[tbp]
\centering
\footnotesize

\caption{\textbf{Detailed 3D simulation results on Adroit, DexArt, and MetaWorld.} Results are averaged over 3 random seeds and reported as mean $\pm$ standard deviation. The best result in each column is highlighted in \colorbox{yellow!25}{\textbf{bold}}.} \vspace{0.5em} 

\setlength{\tabcolsep}{5pt} \label{tab:app_sim_3d_1} \begin{tabular}{l | c | c c c | c c c c} \toprule \multirow{2}{*}{Method} & \multirow{2}{*}{NFE} & \multicolumn{3}{c|}{\textbf{Adroit}} & \multicolumn{4}{c}{\textbf{DexArt}} \\ & & Hammer & Door & Pen & Laptop & Faucet & Bucket & Toilet \\ \midrule DP3 & 100 & \cellcolor{yellow!25}\textbf{100$\pm$0} & 70$\pm$1 & 67$\pm$1 & \cellcolor{yellow!25}\textbf{94$\pm$1} & 44$\pm$1 & 34$\pm$1 & 82$\pm$1 \\ DP3 & 10 & 98$\pm$2 & 66$\pm$4 & 64$\pm$4 & 91$\pm$1 & 43$\pm$1 & 35$\pm$1 & 77$\pm$1 \\ FM Policy & 100 & \cellcolor{yellow!25}\textbf{100$\pm$0} & 76$\pm$1 & 74$\pm$1 & 92$\pm$5 & 39$\pm$4 & 32$\pm$3 & \cellcolor{yellow!25}\textbf{83$\pm$4} \\ FM Policy & 10 & \cellcolor{yellow!25}\textbf{100$\pm$0} & 72$\pm$3 & 69$\pm$3 & 92$\pm$3 & 37$\pm$5 & 34$\pm$5 & 80$\pm$7 \\ CP & 1 & 98$\pm$3 & 67$\pm$5 & 65$\pm$5 & 87$\pm$3 & 44$\pm$4 & 29$\pm$4 & 82$\pm$4 \\ OneDP & 1 & 99$\pm$1 & 67$\pm$2 & 65$\pm$2 & 89$\pm$4 & 45$\pm$3 & 33$\pm$6 & \cellcolor{yellow!25}\textbf{83$\pm$4} \\ MP1 & 1 & \cellcolor{yellow!25}\textbf{100$\pm$0} & 74$\pm$3 & 72$\pm$3 & 88$\pm$5 & 42$\pm$1 & 33$\pm$0 & 82$\pm$2 \\ OFP & 1 & \cellcolor{yellow!25}\textbf{100$\pm$0} & \cellcolor{yellow!25}\textbf{79$\pm$7} & \cellcolor{yellow!25}\textbf{76$\pm$7} & 92$\pm$3 & \cellcolor{yellow!25}\textbf{46$\pm$4} & \cellcolor{yellow!25}\textbf{39$\pm$0} & 81$\pm$3 \\ \bottomrule \end{tabular} \vspace{1em} 

\setlength{\tabcolsep}{4pt}
\label{tab:app_sim_3d_2}
\begin{tabular}{l | c | c c c c c c}
\toprule
\multirow{3}{*}{Method} & \multirow{3}{*}{NFE} & \multicolumn{6}{c}{\textbf{MetaWorld Easy}} \\
 & & Button & Button Press & Button Press & Button & Coffee & Dial \\
 & & Press & Topdown & Topdown Wall & Press Wall & Button & Turn \\
\midrule
DP3 & 100 & 78$\pm$10 & \cellcolor{yellow!25}\textbf{100$\pm$0} & 98$\pm$2 & 82$\pm$8 & 81$\pm$13 & 79$\pm$14 \\
DP3 & 10 & 75$\pm$7 & \cellcolor{yellow!25}\textbf{100$\pm$0} & 96$\pm$3 & 83$\pm$6 & 81$\pm$5 & 70$\pm$4 \\
FM Policy & 100 & 50$\pm$14 & \cellcolor{yellow!25}\textbf{100$\pm$0} & \cellcolor{yellow!25}\textbf{100$\pm$0} & 53$\pm$2 & 76$\pm$3 & 68$\pm$11 \\
FM Policy & 10 & 46$\pm$12 & \cellcolor{yellow!25}\textbf{100$\pm$0} & \cellcolor{yellow!25}\textbf{100$\pm$0} & 47$\pm$11 & 73$\pm$7 & 60$\pm$9 \\
CP & 1 & 60$\pm$4 & \cellcolor{yellow!25}\textbf{100$\pm$0} & 94$\pm$3 & 62$\pm$7 & 66$\pm$4 & 54$\pm$8 \\
OneDP & 1 & 74$\pm$5 & \cellcolor{yellow!25}\textbf{100$\pm$0} & \cellcolor{yellow!25}\textbf{100$\pm$0} & 76$\pm$8 & 81$\pm$5 & 66$\pm$5 \\
MP1 & 1 & 61$\pm$3 & \cellcolor{yellow!25}\textbf{100$\pm$0} & \cellcolor{yellow!25}\textbf{100$\pm$0} & 62$\pm$9 & 67$\pm$3 & 53$\pm$13 \\
OFP & 1 & \cellcolor{yellow!25}\textbf{86$\pm$7} & \cellcolor{yellow!25}\textbf{100$\pm$0} & \cellcolor{yellow!25}\textbf{100$\pm$0} & \cellcolor{yellow!25}\textbf{87$\pm$8} & \cellcolor{yellow!25}\textbf{92$\pm$6} & \cellcolor{yellow!25}\textbf{89$\pm$4} \\
\bottomrule
\end{tabular}

\vspace{1em}

\setlength{\tabcolsep}{4pt}
\label{tab:app_sim_3d_3}
\begin{tabular}{l | c | c c c c c c}
\toprule
\multirow{3}{*}{Method} & \multirow{3}{*}{NFE} & \multicolumn{6}{c}{\textbf{MetaWorld Easy}} \\
 & & Door & Door & Door & Door & Drawer & Drawer \\
 & & Close & Lock & Open & Unlock & Close & Open \\
\midrule
DP3 & 100 & \cellcolor{yellow!25}\textbf{100$\pm$0} & 49$\pm$20 & \cellcolor{yellow!25}\textbf{100$\pm$0} & \cellcolor{yellow!25}\textbf{100$\pm$0} & 73$\pm$15 & \cellcolor{yellow!25}\textbf{100$\pm$0} \\
DP3 & 10 & \cellcolor{yellow!25}\textbf{100$\pm$0} & 38$\pm$9 & \cellcolor{yellow!25}\textbf{100$\pm$0} & \cellcolor{yellow!25}\textbf{100$\pm$0} & 76$\pm$6 & \cellcolor{yellow!25}\textbf{100$\pm$0} \\
FM Policy & 100 & \cellcolor{yellow!25}\textbf{100$\pm$0} & 23$\pm$13 & \cellcolor{yellow!25}\textbf{100$\pm$0} & \cellcolor{yellow!25}\textbf{100$\pm$0} & 45$\pm$11 & \cellcolor{yellow!25}\textbf{100$\pm$0} \\
FM Policy & 10 & \cellcolor{yellow!25}\textbf{100$\pm$0} & 31$\pm$18 & \cellcolor{yellow!25}\textbf{100$\pm$0} & \cellcolor{yellow!25}\textbf{100$\pm$0} & 41$\pm$15 & \cellcolor{yellow!25}\textbf{100$\pm$0} \\
CP & 1 & \cellcolor{yellow!25}\textbf{100$\pm$0} & 25$\pm$2 & \cellcolor{yellow!25}\textbf{100$\pm$0} & 90$\pm$5 & 54$\pm$6 & \cellcolor{yellow!25}\textbf{100$\pm$0} \\
OneDP & 1 & \cellcolor{yellow!25}\textbf{100$\pm$0} & 29$\pm$1 & \cellcolor{yellow!25}\textbf{100$\pm$0} & \cellcolor{yellow!25}\textbf{100$\pm$0} & 69$\pm$6 & \cellcolor{yellow!25}\textbf{100$\pm$0} \\
MP1 & 1 & \cellcolor{yellow!25}\textbf{100$\pm$0} & 21$\pm$17 & \cellcolor{yellow!25}\textbf{100$\pm$0} & \cellcolor{yellow!25}\textbf{100$\pm$0} & 56$\pm$5 & \cellcolor{yellow!25}\textbf{100$\pm$0} \\
OFP & 1 & \cellcolor{yellow!25}\textbf{100$\pm$0} & \cellcolor{yellow!25}\textbf{61$\pm$12} & \cellcolor{yellow!25}\textbf{100$\pm$0} & \cellcolor{yellow!25}\textbf{100$\pm$0} & \cellcolor{yellow!25}\textbf{81$\pm$13} & \cellcolor{yellow!25}\textbf{100$\pm$0} \\
\bottomrule
\end{tabular}

\vspace{1em}

\setlength{\tabcolsep}{4pt}
\label{tab:app_sim_3d_4}
\begin{tabular}{l | c | c c c c c c}
\toprule
\multirow{3}{*}{Method} & \multirow{3}{*}{NFE} & \multicolumn{6}{c}{\textbf{MetaWorld Easy}} \\
 & & Faucet & Faucet & Handle & Handle & Handle & Handle \\
 & & Close & Open & Press & Pull & Press Side & Pull Side \\
\midrule
DP3 & 100 & \cellcolor{yellow!25}\textbf{100$\pm$0} & 80$\pm$15 & \cellcolor{yellow!25}\textbf{100$\pm$0} & 47$\pm$21 & 54$\pm$7 & \cellcolor{yellow!25}\textbf{84$\pm$12} \\
DP3 & 10 & \cellcolor{yellow!25}\textbf{100$\pm$0} & 75$\pm$11 & \cellcolor{yellow!25}\textbf{100$\pm$0} & 42$\pm$6 & 57$\pm$5 & 80$\pm$6 \\
FM Policy & 100 & \cellcolor{yellow!25}\textbf{100$\pm$0} & 77$\pm$9 & \cellcolor{yellow!25}\textbf{100$\pm$0} & 30$\pm$9 & 55$\pm$12 & 36$\pm$9 \\
FM Policy & 10 & \cellcolor{yellow!25}\textbf{100$\pm$0} & 72$\pm$10 & \cellcolor{yellow!25}\textbf{100$\pm$0} & 30$\pm$11 & 51$\pm$10 & 35$\pm$19 \\
CP & 1 & \cellcolor{yellow!25}\textbf{100$\pm$0} & 64$\pm$5 & \cellcolor{yellow!25}\textbf{100$\pm$0} & 21$\pm$4 & 45$\pm$5 & 48$\pm$6 \\
OneDP & 1 & \cellcolor{yellow!25}\textbf{100$\pm$0} & 79$\pm$4 & \cellcolor{yellow!25}\textbf{100$\pm$0} & 33$\pm$5 & 50$\pm$8 & 70$\pm$4 \\
MP1 & 1 & \cellcolor{yellow!25}\textbf{100$\pm$0} & 65$\pm$5 & \cellcolor{yellow!25}\textbf{100$\pm$0} & 34$\pm$9 & 45$\pm$13 & 59$\pm$16 \\
OFP & 1 & \cellcolor{yellow!25}\textbf{100$\pm$0} & \cellcolor{yellow!25}\textbf{88$\pm$7} & \cellcolor{yellow!25}\textbf{100$\pm$0} & \cellcolor{yellow!25}\textbf{63$\pm$13} & \cellcolor{yellow!25}\textbf{91$\pm$6} & 69$\pm$10 \\
\bottomrule
\end{tabular}

\vspace{1em}

\setlength{\tabcolsep}{4pt}
\label{tab:app_sim_3d_5}
\begin{tabular}{l | c | c c c c c c}
\toprule
\multirow{3}{*}{Method} & \multirow{3}{*}{NFE} & \multicolumn{6}{c}{\textbf{MetaWorld Easy}} \\
 & & Lever & Plate & Plate Slide & Plate Slide & Plate Slide & \multirow{2}{*}{Reach} \\
 & & Pull & Slide & Back & Back Side & Side & \\
\midrule
DP3 & 100 & \cellcolor{yellow!25}\textbf{78$\pm$14} & 43$\pm$21 & 90$\pm$13 & \cellcolor{yellow!25}\textbf{100$\pm$0} & 87$\pm$9 & \cellcolor{yellow!25}\textbf{77$\pm$11} \\
DP3 & 10 & 66$\pm$3 & 39$\pm$10 & \cellcolor{yellow!25}\textbf{99$\pm$0} & \cellcolor{yellow!25}\textbf{100$\pm$0} & 87$\pm$5 & 69$\pm$5 \\
FM Policy & 100 & 37$\pm$6 & 24$\pm$10 & 69$\pm$9 & \cellcolor{yellow!25}\textbf{100$\pm$0} & 60$\pm$10 & 49$\pm$11 \\
FM Policy & 10 & 32$\pm$17 & 24$\pm$7 & 63$\pm$18 & \cellcolor{yellow!25}\textbf{100$\pm$0} & 58$\pm$14 & 45$\pm$9 \\
CP & 1 & 43$\pm$6 & 19$\pm$2 & 79$\pm$4 & 96$\pm$4 & 73$\pm$7 & 40$\pm$1 \\
OneDP & 1 & 58$\pm$3 & 27$\pm$5 & 95$\pm$1 & \cellcolor{yellow!25}\textbf{100$\pm$0} & 89$\pm$6 & 54$\pm$3 \\
MP1 & 1 & 46$\pm$14 & 28$\pm$15 & 77$\pm$8 & \cellcolor{yellow!25}\textbf{100$\pm$0} & 73$\pm$5 & 43$\pm$13 \\
OFP & 1 & 74$\pm$12 & \cellcolor{yellow!25}\textbf{57$\pm$20} & 95$\pm$5 & \cellcolor{yellow!25}\textbf{100$\pm$0} & \cellcolor{yellow!25}\textbf{94$\pm$3} & 71$\pm$13 \\
\bottomrule
\end{tabular}

\end{table*}

\begin{table*}[tbp]
\centering
\scriptsize

\addtocounter{table}{-1}
\caption{\textbf{(Continued)} Detailed 3D simulation results on Adroit, DexArt, and MetaWorld.}
\vspace{0.5em}
\footnotesize

\setlength{\tabcolsep}{4pt}
\label{tab:app_sim_3d_6}
\begin{tabular}{l | c | c c c c | c c}
\toprule
\multirow{3}{*}{Method} & \multirow{3}{*}{NFE} & \multicolumn{4}{c|}{\textbf{MetaWorld Easy}} & \multicolumn{2}{c}{\textbf{MetaWorld Medium}} \\
 & & Reach & Window & Window & Peg Unplug & Basket & Bin \\
 & & Wall & Close & Open & Side & ball & Picking \\
\midrule
DP3 & 100 & 62$\pm$14 & \cellcolor{yellow!25}\textbf{100$\pm$0} & 87$\pm$7 & 90$\pm$11 & \cellcolor{yellow!25}\textbf{100$\pm$0} & 19$\pm$3 \\
DP3 & 10 & 57$\pm$2 & \cellcolor{yellow!25}\textbf{100$\pm$0} & 81$\pm$10 & 91$\pm$3 & \cellcolor{yellow!25}\textbf{100$\pm$0} & 17$\pm$5 \\
FM Policy & 100 & 24$\pm$8 & \cellcolor{yellow!25}\textbf{100$\pm$0} & 84$\pm$7 & 67$\pm$15 & \cellcolor{yellow!25}\textbf{100$\pm$0} & 17$\pm$6 \\
FM Policy & 10 & 20$\pm$11 & \cellcolor{yellow!25}\textbf{100$\pm$0} & 77$\pm$16 & 62$\pm$14 & \cellcolor{yellow!25}\textbf{100$\pm$0} & 16$\pm$9 \\
CP & 1 & 30$\pm$7 & \cellcolor{yellow!25}\textbf{100$\pm$0} & 69$\pm$4 & 79$\pm$8 & 73$\pm$9 & 7$\pm$3 \\
OneDP & 1 & 44$\pm$5 & \cellcolor{yellow!25}\textbf{100$\pm$0} & 85$\pm$6 & 95$\pm$1 & \cellcolor{yellow!25}\textbf{100$\pm$0} & 13$\pm$5 \\
MP1 & 1 & 35$\pm$10 & \cellcolor{yellow!25}\textbf{100$\pm$0} & 69$\pm$13 & 78$\pm$8 & 78$\pm$11 & 14$\pm$6 \\
OFP & 1 & \cellcolor{yellow!25}\textbf{72$\pm$8} & \cellcolor{yellow!25}\textbf{100$\pm$0} & \cellcolor{yellow!25}\textbf{94$\pm$3} & \cellcolor{yellow!25}\textbf{96$\pm$2} & \cellcolor{yellow!25}\textbf{100$\pm$0} & \cellcolor{yellow!25}\textbf{29$\pm$11} \\
\bottomrule
\end{tabular}

\vspace{1em}

\setlength{\tabcolsep}{3.8pt}
\label{tab:app_sim_3d_7}
\begin{tabular}{l | c | c c c c c c c}
\toprule
\multirow{3}{*}{Method} & \multirow{3}{*}{NFE} & \multicolumn{7}{c}{\textbf{MetaWorld Medium}} \\
 & & Box & Coffee & Coffee & \multirow{2}{*}{Hammer} & Peg Insert & Push & \multirow{2}{*}{Soccer} \\
 & & Close & Pull & Push & & Side & Wall & \\
\midrule
DP3 & 100 & 21$\pm$3 & 43$\pm$11 & 58$\pm$5 & 69$\pm$7 & 53$\pm$2 & 51$\pm$7 & 16$\pm$7 \\
DP3 & 10 & 20$\pm$1 & 43$\pm$3 & 55$\pm$2 & 70$\pm$3 & 55$\pm$3 & 53$\pm$6 & 13$\pm$2 \\
FM Policy & 100 & 18$\pm$3 & 56$\pm$7 & 45$\pm$6 & 63$\pm$1 & 46$\pm$5 & 39$\pm$10 & 11$\pm$4 \\
FM Policy & 10 & 14$\pm$6 & 53$\pm$11 & 47$\pm$6 & 61$\pm$9 & 47$\pm$10 & 36$\pm$13 & 12$\pm$5 \\
CP & 1 & 6$\pm$2 & 37$\pm$6 & 39$\pm$5 & 54$\pm$6 & 39$\pm$3 & 37$\pm$8 & 6$\pm$1 \\
OneDP & 1 & 12$\pm$6 & 41$\pm$12 & 42$\pm$12 & 57$\pm$15 & 43$\pm$11 & 44$\pm$19 & 14$\pm$7 \\
MP1 & 1 & 13$\pm$4 & 43$\pm$5 & 45$\pm$5 & 60$\pm$6 & 45$\pm$10 & 43$\pm$7 & 9$\pm$3 \\
OFP & 1 & \cellcolor{yellow!25}\textbf{27$\pm$6} & \cellcolor{yellow!25}\textbf{57$\pm$10} & \cellcolor{yellow!25}\textbf{59$\pm$5} & \cellcolor{yellow!25}\textbf{73$\pm$3} & \cellcolor{yellow!25}\textbf{59$\pm$7} & \cellcolor{yellow!25}\textbf{57$\pm$5} & \cellcolor{yellow!25}\textbf{20$\pm$7} \\
\bottomrule
\end{tabular}

\vspace{1em}

\setlength{\tabcolsep}{4pt}
\label{tab:app_sim_3d_8}
\begin{tabular}{l | c | c c | c c c c c}
\toprule
\multirow{3}{*}{Method} & \multirow{3}{*}{NFE} & \multicolumn{2}{c|}{\textbf{MetaWorld Medium}} & \multicolumn{5}{c}{\textbf{MetaWorld Hard}}\\
 & & \multirow{2}{*}{Sweep} & Sweep & \multirow{2}{*}{Assembly} & Hand & Pick Out & Pick & Push \\
 & & & Into & & Insert & of Hole & Place & \\
\midrule
DP3 & 100 & 68$\pm$10 & 15$\pm$4 & 73$\pm$10 & 21$\pm$11 & 22$\pm$13 & 26$\pm$10 & 50$\pm$6 \\
DP3 & 10 & 66$\pm$7 & 17$\pm$3 & 72$\pm$7 & 18$\pm$4 & 20$\pm$10 & 25$\pm$14 & 47$\pm$11 \\
FM Policy & 100 & 58$\pm$7 & 12$\pm$8 & \cellcolor{yellow!25}\textbf{75$\pm$7} & 22$\pm$1 & 26$\pm$5 & \cellcolor{yellow!25}\textbf{46$\pm$10} & 49$\pm$5 \\
FM Policy & 10 & 57$\pm$9 & 12$\pm$6 & 69$\pm$6 & 19$\pm$7 & \cellcolor{yellow!25}\textbf{26$\pm$4} & 41$\pm$14 & 47$\pm$10 \\
CP & 1 & 50$\pm$8 & 6$\pm$2 & 63$\pm$9 & 11$\pm$6 & 14$\pm$8 & 22$\pm$7 & 35$\pm$7 \\
OneDP & 1 & 54$\pm$11 & 16$\pm$8 & 72$\pm$4 & 17$\pm$9 & 19$\pm$2 & 38$\pm$9 & 48$\pm$5 \\
MP1 & 1 & 56$\pm$15 & 11$\pm$4 & 63$\pm$6 & 17$\pm$10 & 19$\pm$11 & 28$\pm$13 & 45$\pm$8 \\
OFP & 1 & \cellcolor{yellow!25}\textbf{70$\pm$7} & \cellcolor{yellow!25}\textbf{24$\pm$12} & 74$\pm$5 & \cellcolor{yellow!25}\textbf{22$\pm$2} & 23$\pm$2 & 42$\pm$1 & \cellcolor{yellow!25}\textbf{55$\pm$6} \\
\bottomrule
\end{tabular}

\vspace{1em}

\setlength{\tabcolsep}{4pt}
\label{tab:app_sim_3d_9}
\begin{tabular}{l | c | c c c c c | c}
\toprule
\multirow{3}{*}{Method} & \multirow{3}{*}{NFE} & \multicolumn{5}{c|}{\textbf{MetaWorld Very Hard}} & \multirow{3}{*}{Average} \\
 & & Shelf & \multirow{2}{*}{Disassemble} & Stick & Stick & Pick Place & \\
 & & Place & & Pull & Push & Wall & \\
\midrule
DP3 & 100 & 22$\pm$5 & 45$\pm$4 & 24$\pm$1 & 74$\pm$8 & 40$\pm$3 & 66.4$\pm$5.7 \\
DP3 & 10 & 21$\pm$9 & 52$\pm$7 & 28$\pm$8 & 80$\pm$7 & 47$\pm$8 & 65.2$\pm$2.8 \\
FM Policy & 100 & \cellcolor{yellow!25}\textbf{26$\pm$6} & 54$\pm$2 & 36$\pm$5 & 82$\pm$8 & 41$\pm$5 & 59.8$\pm$4.6 \\
FM Policy & 10 & 25$\pm$9 & 52$\pm$6 & \cellcolor{yellow!25}\textbf{36$\pm$4} & 79$\pm$14 & 42$\pm$7 & 57.9$\pm$7.0 \\
CP & 1 & 12$\pm$2 & 42$\pm$4 & 21$\pm$1 & 69$\pm$10 & 36$\pm$6 & 54.7$\pm$2.9 \\
OneDP & 1 & 16$\pm$5 & 48$\pm$6 & 26$\pm$6 & 76$\pm$8 & 43$\pm$5 & 62.4$\pm$3.0 \\
MP1 & 1 & 8$\pm$3 & 38$\pm$7 & 18$\pm$4 & 64$\pm$9 & 33$\pm$7 & 57.4$\pm$5.8 \\
OFP & 1 & 23$\pm$2 & \cellcolor{yellow!25}\textbf{56$\pm$2} & 34$\pm$5 & \cellcolor{yellow!25}\textbf{83$\pm$6} & \cellcolor{yellow!25}\textbf{50$\pm$7} & \cellcolor{yellow!25}\textbf{71.6$\pm$4.1} \\
\bottomrule
\end{tabular}

\end{table*}

\section{Ablation Studies}
\label{app:ablation}
To rigorously evaluate the individual contributions of our core design mechanisms and understand their sensitivity to hyperparameter variations, we present a comprehensive ablation study. All evaluations are conducted across 7 simulation tasks from the Adroit and DexArt benchmarks in the 3D pointcloud setting. The policy is trained using 10 demonstrations per Adroit task and 100 demonstrations per DexArt task, utilizing a batch size of 64 over 3000 epochs.\subsection{Component Analysis}We first dissect the distinct roles of Self-Consistency Training (SCT), Self-Guided Regularization (SGR), and the Warm-Start initialization prior. The comparative results are detailed in \cref{tab:ablation_components}.

\begin{table}[t]
\centering
\footnotesize
\setlength{\tabcolsep}{5pt}
\caption{\textbf{Detailed $\pi_{0.5}$ integration results on RoboTwin 2.0.} All methods are evaluated on 100 randomized test trials.}
\label{tab:app_robotwin_full}
\begin{tabular}{l | c | c c c c}
\toprule
\multirow{2}{*}{Method} & \multirow{2}{*}{NFE} & Adjust & Beat Block & Handover & Place Empty \\
 & & Bottle & Hammer & Mic & Cup \\
\midrule
$\pi_{0.5}$ & 10 & 96.0$\pm$4.0 & 88.3$\pm$3.8 & 90.7$\pm$1.2 & 91.0$\pm$3.0 \\
CFM & 1 & 85.3$\pm$3.1 & 82.0$\pm$3.6 & 74.3$\pm$1.2 & 74.7$\pm$2.9 \\
Shortcut & 1 & 89.3$\pm$2.3 & 78.0$\pm$3.6 & 75.0$\pm$3.0 & 81.3$\pm$2.5 \\
iMF & 1 & 92.7$\pm$2.3 & 84.3$\pm$2.5 & 82.7$\pm$2.9 & 87.0$\pm$3.5 \\
OFP & 1 & \cellcolor{yellow!25}\textbf{99.0$\pm$1.0} & \cellcolor{yellow!25}\textbf{91.7$\pm$2.5} & \cellcolor{yellow!25}\textbf{94.3$\pm$3.8} & \cellcolor{yellow!25}\textbf{93.7$\pm$1.5} \\
\bottomrule
\end{tabular}
\end{table}

\begin{table*}[h]
\centering
\footnotesize % Using \small instead of \resizebox prevents the font from scaling up
\caption{\textbf{Component analysis of OFP.} SCT: Self-Consistency Training. SGR: Self-Guided Regularization.}
\label{tab:ablation_components}
\begin{tabular}{l | c | c c c c c c c | c}
\toprule
Method & NFE & Door & Hammer & Pen & Bucket & Faucet & Laptop & Toilet & Average \\
\midrule
\multirow{2}{*}{SCT} 
& 1 & 72$\pm$3 & \cellcolor{yellow!25}\textbf{100$\pm$0} & 68$\pm$5 & 33$\pm$3 & 42$\pm$1 & 89$\pm$1 & 77$\pm$1 & 68.7 \\
& 4 & 82$\pm$1 & \cellcolor{yellow!25}\textbf{100$\pm$0} & 72$\pm$2 & 36$\pm$2 & 44$\pm$3 & 90$\pm$0 & 82$\pm$1 & 72.3 \\
\midrule
\multirow{2}{*}{SCT+SGR}
& 1 & 77$\pm$4 & \cellcolor{yellow!25}\textbf{100$\pm$0} & 73$\pm$3 & 37$\pm$1 & 44$\pm$0 & 90$\pm$1 & 79$\pm$2 & 71.4 \\
& 4 & 82$\pm$3 & \cellcolor{yellow!25}\textbf{100$\pm$0} & 73$\pm$1 & 39$\pm$3 & 43$\pm$1 & 90$\pm$2 & 81$\pm$0 & 72.6 \\
\midrule
\multirow{2}{*}{\textbf{OFP}} 
& 1 & 79$\pm$7 & \cellcolor{yellow!25}\textbf{100$\pm$0} & 76$\pm$7 & 39$\pm$0 & 46$\pm$4 & \cellcolor{yellow!25}\textbf{92$\pm$3} & 81$\pm$3 & 73.3 \\
& 4 & \cellcolor{yellow!25}\textbf{86$\pm$2} & \cellcolor{yellow!25}\textbf{100$\pm$0} & \cellcolor{yellow!25}\textbf{79$\pm$1} & \cellcolor{yellow!25}\textbf{43$\pm$2} & \cellcolor{yellow!25}\textbf{47$\pm$1} & \cellcolor{yellow!25}\textbf{92$\pm$4} & \cellcolor{yellow!25}\textbf{83$\pm$1} & \cellcolor{yellow!25}\textbf{75.7} \\
\bottomrule
\end{tabular}
\end{table*}

The data clearly illustrates the fundamental limitation of relying solely on pure trajectory-matching distillation. When training exclusively with SCT, the policy learns robust temporal coherence, achieving a strong average success rate of 72.3\% at NFE=4. However, its one-step generation capacity lags significantly, yielding only 68.7\%. This performance gap occurs because consistency training aligns the endpoints of the flow trajectory but struggles to enforce precise alignment with the high-density modes of the target data distribution. As a result, one-step predictions often average over multiple valid modes, lacking the sharpness required for precise manipulation. 

Integrating SGR effectively neutralizes this limitation. By leveraging classifier-free guidance as an explicit corrective signal, SGR repels single-step predictions away from unconditional noise and drives them directly toward expert modes. This targeted regularization elevates the one-step success rate to 71.4\%, an increase of nearly 3 percentage points. Furthermore, sharpening the one-step prediction provides more accurate local targets for multi-step integration, pushing the NFE=4 performance to 72.6\%. This demonstrates that SCT and SGR are highly complementary: SCT smooths the flow trajectory across time, while SGR ensures the final prediction lands precisely on the data manifold.

Incorporating the training-free Warm-Start mechanism during inference yields substantial additional gains, raising the success rate to 73.3\% at NFE=1 and 75.7\% at NFE=4. By constructing an initialization prior from the temporally shifted unexecuted actions, Warm-Start inherently positions the starting state much closer to the target distribution. This reduces the transport distance the generative model must traverse in a single step, reliably amplifying control smoothness and precision.

\subsection{Hyperparameter Analysis}
To optimize system efficiency and stability, we explore the selection and design of hyperparameters.
\begin{table*}[tb]
\centering
\captionsetup[subtable]{font=small}
\caption{\textbf{Hyperparameter analysis.} All results reflect the average success rate across the 7 tasks in Adroit and DexArt.}
\label{tab:hyperparameters}
% \vspace{5pt}

% --- Top Row ---
\begin{subtable}[t]{0.48\textwidth}
\centering
\footnotesize
\caption{\textbf{Sub-batch training ratios.} Impact of data fraction allocation ($p_{sc}$, $p_{sg}$).}
% \vspace{2pt}
\label{tab:sub_batch}
\begin{tabular}{c c | c}
\toprule
$p_{sc}$ & $p_{sg}$ & SR (\%) \\
\midrule
20\% & 10\% & \cellcolor{yellow!25}\textbf{73.3} \\
20\% & 20\% & 72.2 \\
50\% & 10\% & 71.7 \\
50\% & 20\% & 71.4 \\
\bottomrule
\end{tabular}
\end{subtable}\hfill
\begin{subtable}[t]{0.48\textwidth}
\centering
\footnotesize
\caption{\textbf{Time samplers.} Performance under various interval sampling distributions for $t$ and $dt$.}
% \vspace{2pt}
\label{tab:time_sampler}
\begin{tabular}{l l | c}
\toprule
$t$ sampler & $dt$ sampler & SR (\%) \\
\midrule
Uniform(0, 1) & Uniform(0, 1) & 69.9 \\
Uniform(0, 1) & LogNorm(-0.2, 1) & 71.7 \\
Uniform(0, 1) & LogNorm(-0.4, 1) & 72.3 \\
Beta(1, 1.5) & LogNorm(-0.2, 1) & \cellcolor{yellow!25}\textbf{73.3} \\
Beta(1, 1.5) & LogNorm(-0.4, 1) & 72.2 \\
\bottomrule
\end{tabular}
\end{subtable}

% \vspace{15pt} % Adds vertical space to separate the two rows neatly

\vspace{6pt}
% --- Bottom Row ---
\begin{subtable}[t]{0.48\textwidth}
\centering
\footnotesize
\caption{\textbf{Guidance scale.} The effect of the SGR loss regularizer weight $\lambda_{sg}$.}
% \vspace{2pt}
\label{tab:lambda_sg}
\begin{tabular}{l | c}
\toprule
$\lambda_{sg}$ & SR (\%) \\
\midrule
0.5  & 68.7 \\
0.1  & 70.4 \\
0.05 & \cellcolor{yellow!25}\textbf{73.3} \\
0.01 & 72.1 \\
\bottomrule
\end{tabular}
\end{subtable}\hfill
\begin{subtable}[t]{0.48\textwidth}
\centering
\footnotesize
\caption{\textbf{Warm-start noise.} Modifying the initialization noise ratio via $t_w$.}
% \vspace{2pt}
\label{tab:warm_start}
\begin{tabular}{l | c}
\toprule
$t_w$ & SR (\%) \\
\midrule
0.75 & 69.5 \\
0.5  & 72.3 \\
0.25 & 72.1 \\
0.15 & \cellcolor{yellow!25}\textbf{73.3} \\
0.0  & 71.4 \\
\bottomrule
\end{tabular}
\end{subtable}

\end{table*}

\textbf{Sub-Batch Allocation.} To mitigate computational overhead, OFP implements a sub-batch training strategy. We assign a proportion $p_{sc}$ of the batch to compute the self-consistency loss ($r \neq t$). The remaining $1 - p_{sc}$ fraction is utilized for standard Flow Matching ($r = t$). From this subset, a proportion $p_{sg}$ is allocated to calculate the SGR loss. \cref{tab:sub_batch} indicates that the optimal configuration is $p_{sc} = 0.2$ and $p_{sg} = 0.1$. Assigning excessive data to consistency or guidance detracts from the foundational flow-matching objective, whereas maintaining smaller, targeted proportions provides optimal regularization without destabilizing the baseline field.

\textbf{Time Sampling Distribution.} Generation quality is sensitive to the sampling distributions of the interval times during consistency training. The policy network takes the current time $t$ and the relative interval $dt$ as conditions, and the interval target at $r = t + dt$. As shown in \cref{tab:time_sampler}, sampling $t$ from a Beta distribution and $dt$ from a LogNormal distribution achieves the highest success rate (73.3\%).

\textbf{Guidance Weight.} The coefficient $\lambda_{sg}$ modulates the strength of the self-guided regularizer relative to the total objective. \cref{tab:lambda_sg} demonstrates that a conservative weight of $\lambda_{sg} = 0.05$ is optimal. Imposing a highly aggressive guidance signal (e.g., 0.5) over-regularizes the generation, forcing the model into mode collapse and actively degrading performance to 68.7\%.

\textbf{Warm-Start Noise Ratio.} Warm-Start initializes the generator by corrupting the unexecuted sequence of a previous action chunk with a specific ratio of Gaussian noise, controlled by the starting time parameter $t_w$. As $t_w \to 0$, the initialization approaches pure noise. \cref{tab:warm_start} establishes that $t_w = 0.15$ yields peak performance. When $t_w$ is excessively small, inference effectively deteriorates into standard zero-shot generation, discarding the powerful temporal prior. Conversely, assigning a large $t_w$ (e.g., 0.75) heavily suppresses necessary generative randomness. This rigid reliance on past predictions restricts the system's ability to correct compounding errors or adapt to large shifts in the immediate action space.

\section{Related Work}
\label{app:related_work}
\subsection{Generative Policies for Robot Control}
Diffusion models initially entered the robotics domain through the planning-as-generation paradigm. Works such as Diffuser \cite{janner2022planning} and Decision Diffuser \cite{ajay2022conditional} conceptualized long-horizon behaviors as trajectory samples generated through gradual denoising, replacing traditional dynamic programming with conditional generation. Subsequently, the focus shifted to directly parameterizing visuomotor policies. Diffusion Policy \cite{chi2025diffusion} modeled action sequence generation as a conditional denoising process, significantly improving the expressiveness of multimodal action distributions and overall training stability. However, its deployment requires iterative multi-step sampling, causing inference latency to scale linearly with the step count. Subsequent research expanded along two primary axes. The first focuses on strengthening spatial representations and structural biases to improve fine manipulation, utilizing approaches like cross-attention (Crossway Diffusion \cite{li2024crossway}), 3D point cloud conditioning (DP3 \cite{ze20243d}, 3D Diffuser Actor \cite{ke20243d}), and key-pose prediction (ChainedDiffuser \cite{xian2023chaineddiffuser}). The second axis addresses the strict latency constraints of closed-loop control by exploring high-throughput paradigms (DiffuserLite \cite{dong2024diffuserlite}).

In parallel, policy learning based on Flow Matching has gained significant traction. Continuous Flow Matching (CFM) has been successfully integrated with affordance representations \cite{zhang2024affordance} and multi-contact whole-body control \cite{rouxel2024flow}. Works like ActionFlow \cite{funk2024actionflow} inject $\mathrm{SE}(3)$ equivariant structures into the velocity field to improve spatial generalization, while AdaFlow \cite{hu2024adaflow} designs adaptive step-size solvers to balance speed and diversity. Within this framework, structured approaches for few-step inference have emerged. For example, FlowPolicy \cite{zhang2025flowpolicy} pursues single-step generation via consistency flow matching, and MP1 \cite{sheng2025mp1} adapts MeanFlow's interval-averaged velocity to generate 3D point cloud-conditioned actions in a single step (1-NFE).

As generative policies scale into massive Vision-Language-Action (VLA) models, the computational cost of iterative sampling becomes a severe bottleneck. Architectures like RDT-1B \cite{liu2024rdt} scale diffusion transformers to a billion parameters for dual-arm manipulation, while the $\pi$ series (including $\pi_{0}$ \cite{black2024pi_0}, $\pi_{0.5}$ \cite{intelligence2025pi_} and $\pi_{0.6}$) \cite{intelligence2025pi} establishes Flow Matching as a highly scalable action generation head for open-world tasks. As model capacity and semantic complexity increase, compressing generation into a single step without degrading control accuracy is a critical requirement for real-world deployment.

\subsection{Distillation and Acceleration}
In visual generation, two dominant paradigms exist for few-step acceleration. Consistency Models \cite{song2023consistency} map points along a probability path to a single origin, establishing the consistency distillation route. Alternatively, Distribution Matching Distillation (DMD) \cite{yin2024improved} uses score differences to align a student model with a teacher's distribution via reverse-KL gradients. Recently, MeanFlow \cite{geng2025mean} enabled from-scratch 1-NFE training by modeling finite-interval average velocity; however, implementing its objective requires complex, high-order Jacobian-Vector Products (JVPs) that destabilize training and consume significant memory.

These acceleration concepts have rapidly transitioned into robotics.  The consistency paradigm was adapted into RL (CPQL \cite{chen2023boosting}) and visuomotor control (Consistency Policy \cite{prasad2024consistency}, ManiCM \cite{lu2024manicm}). Because consistency distillation forces intermediate states to converge to a shared endpoint, its optimization objective is inherently mode-covering. While this enables few-step inference, it frequently causes the policy to average across multimodal action distributions, resulting in smoothed, imprecise actions that fail in fine manipulation tasks.

Conversely, methods like One-Step Diffusion Policy (OneDP) \cite{wang2024one} and SDM Policy \cite{jia2024score} apply DMD-style distribution matching to robotics. This approach is inherently mode-seeking, producing sharper, higher-quality single-step samples. However, it severely limits action diversity. Additionally, it relies heavily on large pre-trained teacher models and auxiliary score networks, which increases training overhead and creates generalization bottlenecks during distribution shifts.

Beyond distillation, works like Streaming Diffusion Policy \cite{hoeg2024streaming} exploit the temporal correlation of receding-horizon control, reusing intermediate states to reduce required denoising steps and maintain high control frequencies.

\textbf{Summary.} Existing few-step acceleration methods force a trade-off between the mode-covering stability of consistency models and the mode-seeking precision of score distillation. OFP unites both signals within a unified, from-scratch self-distillation objective. By doing so, it eliminates the computational overhead of external teachers and the instability of JVP operators. Combined with a warm-start mechanism that reuses cross-chunk priors, OFP delivers a highly stable, single-step inference solution with the precision required for complex robotic control.

%%%%%%%%%%%%%%%%%%%%%%%%%%%%%%%%%%%%%%%%%%%%%%%%%%%%%%%%%%%%

\end{document}